%% file: hmd.tex
\documentclass[10pt,twocolumn,letterpaper]{article}

\usepackage[pagenumbers]{cvpr} 

\usepackage[utf8]{inputenc} 
\usepackage[T1]{fontenc}    
\usepackage{graphicx}
\usepackage{booktabs,longtable}
\usepackage{tabularx}
\usepackage{xcolor}         
\usepackage{mathtools,amssymb,amsthm,amsfonts}
\usepackage{subcaption}
\usepackage[shortlabels,inline]{enumitem}
\setlist{nosep}
\usepackage{listings}
\usepackage[pagebackref,breaklinks,colorlinks]{hyperref}
\usepackage[capitalize]{cleveref}
\crefname{section}{Sec.}{Secs.}
\Crefname{section}{Section}{Sections}
\Crefname{table}{Table}{Tables}
\crefname{table}{Tab.}{Tabs.}

\input{cgmacros.tex}

\DeclareMathOperator{\clip}{clip}
\newcommand{\cwg}[1]{}
\newcommand{\hjk}[1]{}

\def\cvprPaperID{19} 
\def\confName{The 3rd Workshop of Adversarial Machine Learning on Computer Vision: Art of Robustness (in conjunction with CVPR 2023)}
\def\confYear{2023}

\begin{document}

\title{
How many dimensions are required to find an adversarial example?
}

\author{Charles Godfrey,$^1$ Henry Kvinge,$^{1,2}$ Elise Bishoff,$^1$ Myles Mckay,$^{1,3}$\\ Davis Brown,$^1$ Tim Doster,$^1$ and Eleanor Byler$^1$\\
$^1$Pacific Northwest National Laboratory, 
$^2$Department of Mathematics, University of Washington,\\
$^3$Department of Astronomy, University of Washington\\
{\tt\small \{first\}.\{last\}@pnnl.gov}
}
\maketitle

\begin{abstract}
   \input{abstract.tex}
\end{abstract}

\input{body}
{\small
\bibliographystyle{ieee_fullname}
\bibliography{hmd}
}

\clearpage
\appendix
\input{appendix}

\end{document}

%% file: cgmacros.tex
\newcommand{\RR}{\mathbb{R}}

\newcommand{\nrm}[1]{\lvert #1 \rvert}

\newcommand{\cX}{\mathcal{X}}
\newcommand{\cY}{\mathcal{Y}}

\newcommand{\cD}{\mathcal{D}}

\DeclareMathOperator{\success}{success}
\DeclareMathOperator{\Gr}{Gr}
\DeclareMathOperator{\sign}{sign}

\numberwithin{equation}{section}

\theoremstyle{plain}

\newtheorem{lemma}[equation]{Lemma}

\theoremstyle{definition}

\theoremstyle{remark}

%% file: abstract.tex
Past work exploring adversarial vulnerability have focused on situations
where an adversary can perturb all dimensions of model input. On the other hand,
a range of recent works consider the case where either (i) an adversary can
perturb a limited number of input parameters or (ii) a subset of modalities in a
multimodal problem. In both of these cases, adversarial examples are
effectively constrained to a subspace $V$ in the ambient input space $\cX$.
Motivated by this, in this work we investigate how adversarial vulnerability
depends on $\dim(V)$. In particular, we show that the adversarial success of
standard PGD attacks with \(\ell^p\) norm constraints behaves like a monotonically increasing function of \(\epsilon
(\frac{\dim(V)}{\dim \cX})^{\frac{1}{q}}\) where \(\epsilon\) is the perturbation budget and
\(\frac{1}{p} + \frac{1}{q} =1\), provided \( p > 1 \) (the case \(p=1\) presents additional subtleties which we analyze in some detail). This functional form can be easily derived
from a simple toy linear model, and as such our results land further credence to
arguments that adversarial examples are endemic to locally linear models on
high dimensional spaces.

%% file: body.tex
\section{Introduction}


\begin{figure*}[tb]
    \centering
   \begin{subfigure}{0.48\linewidth}
        \centering
        \includegraphics[width=\textwidth]{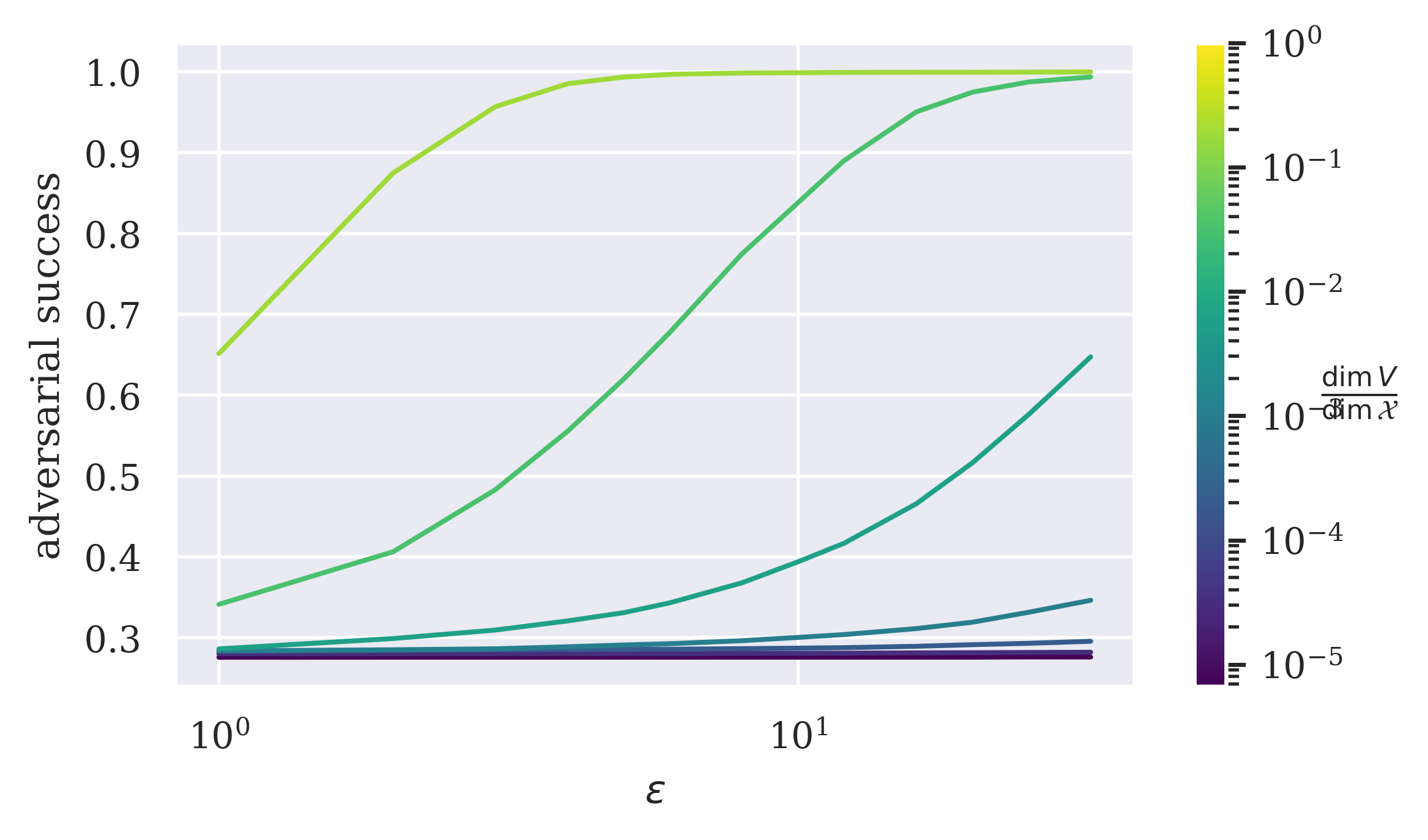}
    \end{subfigure}
    \begin{subfigure}{0.48\linewidth}
        \centering
        \includegraphics[width=\linewidth]{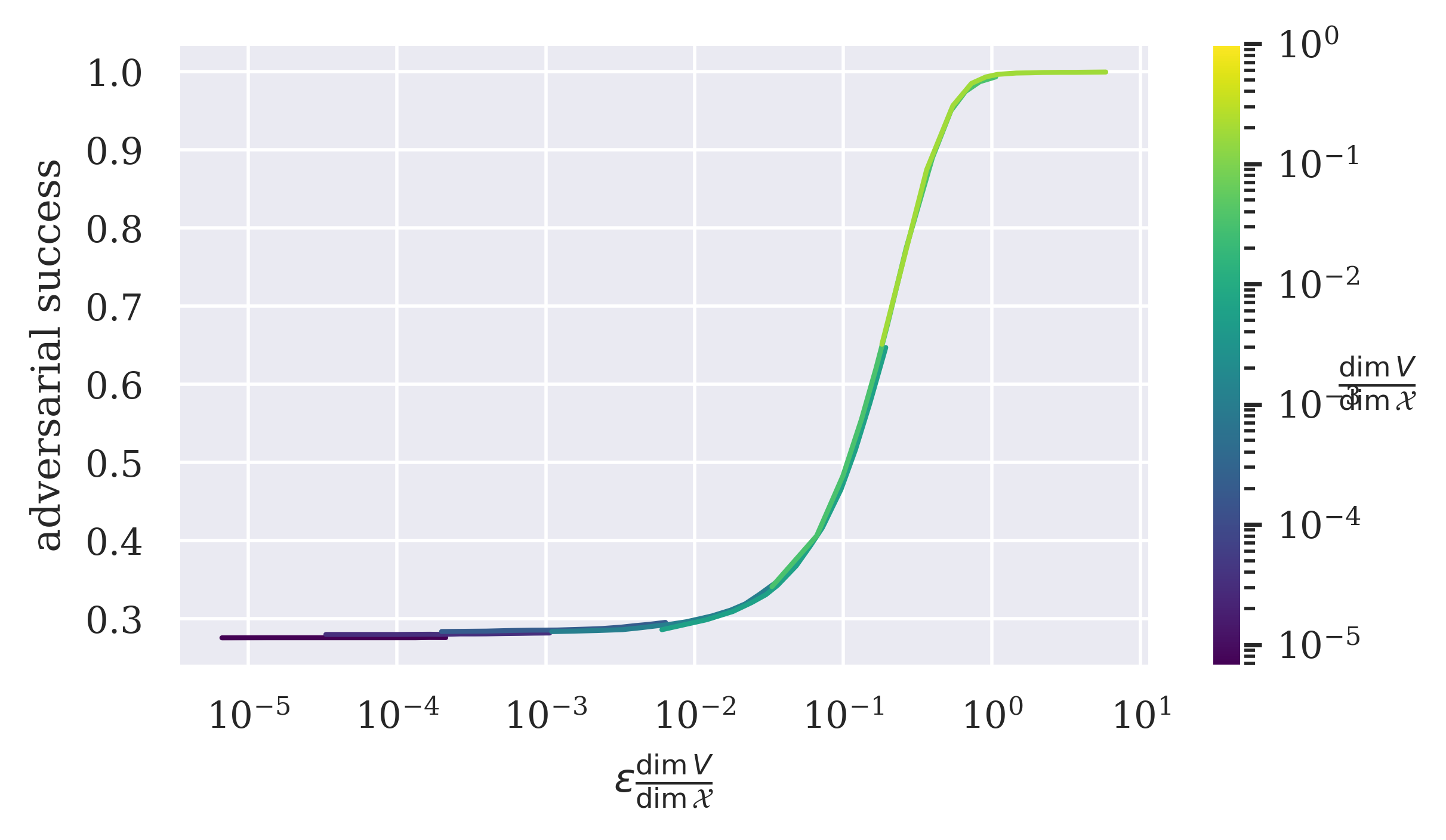}
    \end{subfigure}
   \caption[]{\textbf{Left}: success of PGD adversarial attacks on an ImageNet trained ResNet50, with $\ell^\infty$-norm constraints on perturbation budget, constrained to subspaces \(V \subseteq \cX\) spanned by \(\dim V\) randomly selected standard basis vectors. Adversarial examples are computed for a random subsample of 10,000 datapoints from the ImageNet validations set. The $x$-axis is the $\epsilon$-bound used during example generation and the different colored curves indicate the dimension \(\dim V\) of the subspace to which the examples were constrained to, relative to the dimension \(\dim \cX \) (\(=3\cdot 224^2\)) of the ambient space. When only a small number of dimensions can be perturbed, adversarial examples are challenging to generate even with large $\epsilon$-bounds. \textbf{Right}: these curves become aligned when we reparameterize the $x$-axis by scaling by $\frac{\dim V}{\dim \cX}$.}
   \label{fig:infty}
\end{figure*}

Since they were first identified in \cite{intrigue}, there has been strong sense that a model's vulnerability to adversarial examples is strongly connected to the dimension of its input space. This connection has been mined by a range of works which use it as a perspective with which to explain the prevalence of adversarial examples in certain model types (e.g., computer vision) -- in \cref{sec:rw} we provide a brief synopsis of this research. As deep learning models are applied to more and more safety critical applications, there is also an increasing practical relevance to understanding any general connections between adversarial vulnerability and the properties of a problem. In such settings, a simple statistic that can be easily computed (such as model input dimension) is useful for gauging the general adversarial risk for a proposed deep learning system. 

This is especially true when the proposed system uses less familiar modalities/tasks to which one cannot easily refer to studies in the literature. For example, suppose one needs to evaluate the safety of applying deep learning to the output of a range of different sensors. Past work has considered the ambient dimension in which this data is collected. Should we worry less if a sensor captures a signal as a $50$-dimensional vector rather than a $5,000$-dimensional vector? In this paper we take this line of reasoning a step further and ask how this situation changes when instead of changing the ambient dimension we change the dimension of the subspace in which one is constrained to perturb input. Such a thought experiment has practical relevance. Suppose that of the $500$ input dimensions to our model, we believe that an adversary is only likely to get access to $50$ dimensions (this may happen in multimodal settings where an adversary has much better access to a subset of the modalities). How should we compare this to a situation in which are only able to perturb a fixed $100$-dimensional subspace of the input? How about a $5$-dimensional subspace?

Motivated by this, in this work we revisit the connection between dimension and adversarial vulnerability. 
Unlike most other works in this space, which look at susceptibility to adversarial examples as a function of the number of input dimensions $\dim(\cX)$ alone, we explore model susceptibility to adversarial examples constrained to a subspace $V \subseteq \cX$ as a function of $\dim(V)/\dim(\mathcal{X})$. We find that unsurprisingly, for fixed $\dim(\mathcal{X})$, as $\dim(V)$ decreases average adversarial success rate (ASR) also decreases, though ASR only drops significantly when the quotient $\dim(V)/\dim(\mathcal{X})$ drops below around $10\%$ (see Figure \ref{fig:infty}). In other words, a model remains vulnerable when an adversary is only able to perturb a subset of input dimensions, but as this subset covers an ever smaller fraction of the available dimensions an adversary has to put increasing effort into finding adversarial examples. 

We further study how the adversarial budget $\epsilon$ with respect to the $\ell^p$-norm interacts with $\dim(V)$ and $\dim(\mathcal{X})$. We find that the relationships of $\epsilon$ to ASR for different $\dim(V)$ are nearly identical up to scaling: more specifically, suppose that $C_{V_1}: \mathbb{R} \rightarrow [0,1]$ and $C_{V_2}: \mathbb{R} \rightarrow [0,1]$ map adversarial budget to ASR when adversarial examples are constrained to subspaces $V_1$ and $V_2$ respectively. We find that 
\begin{equation}
\label{eq:rescaling}
    C_1\Big( \Big(\frac{\dim(V_1)}{\dim(\mathcal{X})}\Big)^{1/q}\epsilon\Big) \approx C_1\Big( \Big(\frac{\dim(V_2)}{\dim(\mathcal{X})}\Big)^{1/q}\epsilon\Big)
\end{equation}
where $q$ satisfies $1/p + 1/q = 1$. Emprical results for ImageNet trained
ResNet50 models in the case \(p=\infty \) are displayed in \cref{fig:infty}, and
results for other datsets, models and values of \(p\) can be found in
\cref{sec:exper,sec:add-exper}. \Cref{eq:rescaling} points to a strong relationship between
$\dim(V)$, $p$, and $\epsilon$ that to our knowledge is novel. It further tells
us that risk from adversarial examples can be mitigated by either restricting
the dimensions that data can be manipulated (\(\dim V \)) in or restricting the
amount they can be manipulated before they are noticed (\(\epsilon\)). This
relationship is consistent across values of $\dim(V)/\dim(\mathcal{X})$: if one
wanted to understand the risk of an adversary purturbing data in a \(50
\)-dimensional subspace of a 500-dimensional-input space, one could for example
estimate the success rate of an adversary with access to the entire input space
and extrapolate using \cref{eq:rescaling}.  Finally, we provide a theoretical
backing for our results as well as analyze their implications on common theories
behind the prevalence of adversarial examples in Section \ref{sec:comp-w-thry}.

In summary, our contributions in this paper include the following.
\begin{itemize}
\item We run a range of experiments restricting adversarial examples to a fixed subspace $V$ of input space and explore how the dimension of $V$ impacts adversarial success rate (ASR).
\item Our results show that there are predictable trade-offs between $\epsilon$ and $\dim(V)$. That is, we can scale $\epsilon$ and $\dim(V)$ (dependent on the $\ell^p$-norm used) so that ASR remains fixed.
\item We provide a theoretical basis for our observations and analyze what this says about different theories explaining the existence of adversarial examples.
\end{itemize}

\section{Related work}
\label{sec:rw}

\noindent\textbf{Why do adversarial examples exist?} There have been many proposed explanations for the phenomena of adversarial
examples; we provide an incomplete but representative sample. A number of works such as
\cite{shafahi2018are,curse} present explanations in terms of dimensionality
curses. In \cite{harness} it is argued that adversarial examples are a
side-effect of locally linear behavior of deep learning models, an idea that is
further investigated both theoretically and empirically in
\cite{cubukIntriguingPropertiesAdversarial2017}. This theme also appears in
\cite{bubeckSingleGradientStep2021,bartlettAdversarialExamplesMultiLayer2021},
which (among many other things) prove ReLU networks with multiple layers are
linear on large regions of input space.\\

\noindent\textbf{Adversarial examples and input dimension:} A range of works have looked at the connection between the input dimension of data to a model and the prevalence of adversarial examples. Such works include \cite{pmlr-v97-simon-gabriel19a}, which simplifies the set-up by approximating neural networks with their
gradients, hence reducing the problem to linear classifiers. They vary input dimension
by up-sampling CIFAR10. \cite{shafahi2018are} derives formulae relating
adversarial vulnerability to model input dimension \(\dim \cX\), adversarial
budget \(\epsilon\) (in arbitrary \(\ell^p\) norms, including \(p=0\)) and
notably properties of the data distribution, and carries out experiments varying
input dimension by up-sampling MNIST. \cite{curse} includes theoretical results
of a similar flavor, and also varies input dimension of image datasets by
up-sampling as well as dimension-reducing preprocessing operations like the
singular value decomposition. Unlike our work, none of these considered adversarial examples constrained to subspaces \(V \subset \cX\).\footnote{At
first glance it might seem the SVD preprocessing lands in a proper subspace \(V
\subset \cX\), but it is more accurate to say it decreases the ambient dimension
\(\dim \cX\).}\\

\noindent\textbf{Adversarial examples constrained to subspaces:} There is a continually expanding body of work on adversarial
perturbations constrained to submanifolds of the input space of a model.
\cite{guoLowFrequencyAdversarial2019,YinFourier,Jo2017MeasuringTT,pmlr-v97-rahaman19a,sharmaEffectivenessLowFrequency2019}
all study the vulnerability of neural networks to perturbations constrained to a
subspace corresponding to some Fourier frequency range (for instance, high, low
or intermediate frequencies). \cite{FunAttacks,AdvColor} study vulnerability to
perturbations which modify color curves simultaneously at all locations of an
image.

Among works most in line with the present one, \cite{adv2random} studies the minimal norm perturbation \(\delta \in V \subseteq
\cX\) required to move an input \(x \in \cX\) across a decision boundary of a
classification model
\(f\). They provided theoretical results for linear classifiers (and more
general models in terms of curvature properties of decision boundaries) as well
as empirical results for several image classifiers. The main theorems of
\cite{adv2random} state that the norm of the minimal perturbation \(\delta\)
scales like \(\sqrt{\frac{\dim \cX}{\dim V}}\). However, they do not directly
connect these findings with model error (a.k.a. adversarial success) and their
analysis is limited to the \(\ell^2\) norm (hence their theorems do not
contradict \cref{fig:infty}, which illustrates \(\ell^{\infty}\)
adversarial success). On the other hand, in this work we consider arbitrary $\ell^p$-norm bounds and actually connect $p$ to the rate of growth of adversarial success rate. The work in \cite{adv2random} is also intimately connected with
the DeepFool attack \cite{MoosaviDezfooli2016DeepFoolAS},\footnote{Which does
implicitly indicate the correct scaling for more general \(\ell^p\)-metrics.} as
well as \cite{classrobust,pmlr-v84-franceschi18a}. In contrast, we mostly focus
on PGD attacks due to their universality \cite{madry2018towards} and
prevalence in the adversarial machine learning literature.

A number of works such as \cite{Stutz2019CVPR,geometryadvex,tramer2017space} ask the opposite of our question. Namely, what subspace \(V\subset \cX\) adversarial perturbations tend to lie
in. A consistent finding of \cite{Stutz2019CVPR,geometryadvex} is that in
situations where the data distribution lies on a manifold \(M \subseteq \cX\),
adversarial examples for data points \(x \in M\) tend to lie in the \emph{normal
space} \(NM_x\), whose dimension is the codimension of \(M\) ---
\cite{geometryadvex} observes increasing vulnerability as this codimension
increases. 

Perhaps the work most similar to what we present here is \cite{GuoOrigins},
which investigates the phenomena of low-dimensional adversarial perturbations
with theoretical results and empirical confirmations. Our findings are generally
consistent with theirs, and we build on \cite{GuoOrigins} with an extensive
empirical analysis of \emph{simultaneous} dependence of adversarial success on
\(\dim V\), \(\epsilon\) and the metric under consideration (i.e. \(\ell^{2}\)
or \(\ell^\infty\)). In addition, our experiments involve much larger models and
datasets; we hope the description of our methods in \cref{sec:exp-det}
illustrates that this scaling-up is not trivial.

Finally we note that our results can be tied to a range of studies that look at statistics of adversarial examples with respect to different situations. For example, \cite{cubukIntriguingPropertiesAdversarial2017} studies scaling properties of
adversarial success with respect to the perturbation constraint \(\epsilon\),
obtaining results qualitatively similar to ours along that axis of variation.\\

\section{Adversarial examples and subspaces}

Let \(f: \cX \to \cY\) be a classification model, where \(\cX\) is the space of
input data and \(\cY\) the space of model outputs. We focus on
image classifiers so that \(\cX\) is a space of pixel values (a
hypercube of the form \([0,1]^n\) for some \(n\) depending on image resolution)
and \(\cY = \{1, \dots, K\}\) with \(K\) the number of classes. The models \(f\)
we consider are deep neural networks. By definition an \textbf{adversarial
example} for a data point \((x, y) \in \cX \times \cY\) is a model input \(x' =
x + \delta \in \cX\) such that 
\begin{itemize}
   \item 
   \(f(x') \neq y\) (\(x'\) is misclassified) and
   \item \(d(x', x) < \epsilon\) where \(d\) is some chosen metric and
   \(\epsilon >0\) some chosen constraint (\(x'\) is close to \(x\)).
\end{itemize}
We will measure \textbf{adversarial success} as the probability that \(f(x')\neq
y\); this probability depends on the algorithm used to generate \(x' =
x + \delta \in \cX\), and empirically can be estimated by generating
perturbations for all \(x\) in a validation dataset and computing the error of
\(f\) on the resulting ``adversarial dataset.''\footnote{In particular, we do
not require the model to correctly classify the unperturbed input, i.e. we do
not restrict attention to data points where \(f(x)=y\).} 

Standard methods of generating adversarial examples \cite{intrigue,harness} perturb
model inputs by independently modifying all pixel values, however as early as
\cite{harness} it was observed that sparse perturbations modifying only a subset
of pixel values were also effective. By now there exist a plethora of
adversarial example generation techniques which optimize for perturbations
\(\delta\) constrained to a subspace\footnote{Sometimes, but not always, an
affine linear subspace.} \(V
\subseteq \cX\), in many cases with \(\dim V\) a small fraction of \(\dim \cX\)
--- a common aim of these methods is to modify \(x\) in a way that is
perceptually natural (so that \(x'\) will appear innocuous even to a
human-in-the-loop) while using relatively few parameters. We discuss a
representative sample of such techniques in \cref{sec:rw}. Such widespread
interest in constrained adversarial perturbations \(\delta \in V \subset \cX\)
raises a foundational question:
\begin{equation}
   \label{eq:fun-question}
   \textit{how does adversarial success depend on \(\frac{\dim V}{\dim \cX}\)?}
\end{equation}
In \cref{sec:perturb} we design experiments to measure this dependency for a
variety of families of subspaces \( V \subset \cX\) (including those spanned by
random subsets of pixels or random sets of orthogonal vectors) and metrics
(including \(\ell^{2}\) and \(\ell^\infty\)).\footnote{By definition for any \(p
\geq 1\) the \(\ell^p\)-distance between points \(x, y \in \RR^n\) is \((\sum_i
|x_i - y_i|^p)^\frac{1}{p}\).} We repeatedly find that success of
adversarial attacks constrained in the \(\ell^p\) metric is a function of
\(\epsilon \cdot (\frac{\dim V}{\dim \cX})^{\frac{1}{q}},\) where \(\frac{1}{p}
+ \frac{1}{q} = 1\): this is illustrated in
\cref{fig:infty} and described further in \cref{sec:eps-d-pinv}. 

This experiment serves as a lens through which to investigate two common,
not-necessarily-mutually-exclusive 
explanations for the prevalence of adversarial examples in deep learning ---
these are:
\begin{enumerate}[(i)]
   \item \label{item:curse} \textbf{Adversarial examples are a result of the
   curse of dimensionality:} a deep learning model \(f: \cX \to \cY \)
   subdivides the high dimensional input space \(\cX \) into decision regions
   \(f^{-1}(y) \subseteq \cX\). A variety of well-studied toy models, such as
   binary linear classification of points on a sphere \(S^{n-1} \subset
   \RR^n\), have the property that in high dimensions every \(x \in f^{-1}(y)\)
   lies very close to the boundary of \(f^{-1}(y)\) (for linear classification
   of points on \(S^{n-1}\) this is the statement that ``as \(n \to \infty\), all the volume lies
   near the equator''). 
   \item \label{item:linearity} \textbf{Adversarial examples are a consequence
   of (locally) linear behavior:} At least locally, \(f\) is well approximated
   by an affine linear function \(W x + b\), and for appropriately chosen
   \(\delta\) we can make \(|W \delta|\) large enough to ensure \(f(x + \delta)
   \neq f(x)\). 
\end{enumerate}
These two explanations are more closely related than they might initially
appear. For example, in \cref{item:linearity} the number of coefficients of
\(w\) equals \(\dim \cX\), and as pointed out in \cite{harness} the fast
gradient sign method exploits the fact that when \(\delta = - \sign(w)\), \(w^T
\delta = - \sum_i \nrm{w_i} = - \nrm{w}_1\) which scales with \(\dim \cX\)
(provided the scale of the coefficients \(w_1, \dots, w_{\dim \cX}\) is fixed).
Here the idea is that the number of terms in the sum \( \sum_i \nrm{w_i}\)
is \(\dim \cX\), so if the coefficients \(w_i\) are IID \(E[\sum_i \nrm{w_i}] =
\dim \cX \cdot E[\nrm{w_1}] \) --- in some sense, this is also a curse of
dimensionality. In \cref{sec:comp-w-thry} we compare the
various \emph{theoretically predicted} scaling properties of adversarial success
with respect to \(\epsilon\) and \(\dim V\), lifted from papers arguing for
\cref{item:curse} or \cref{item:linearity}.


\section{Perturbations in random subspaces}
\label{sec:perturb}
Designing an experiment to measure adversarial success with varying \(\dim V\)
and \(\epsilon\) requires making a number  of choices:
\begin{enumerate}[(i)]
   \item \label{item:dist-on-Gr} a distribution of subspaces \(V \subseteq \cX\) to sample from,
   or more technically speaking a probability distribution on a Grassmannian \(\Gr(\dim
   V, \dim \cX)\),
   \item \label{item:metric} a metric \(d\) used to define constraints on perturbations, and
   \item \label{item:algorithm} an adversarial example generation algorithm \(\mathcal{A}\).
\end{enumerate}
To establish a baseline, we consider the case where the distribution of
subspaces is either \emph{uniform} or the distribution obtained by taking \(V\)
to be the span of \(\dim V\) standard basis vectors \(e_i \in \cX\) sampled
uniformly. In our experiments we restrict attention to the \(\ell^p\) metrics for \(p\in \{1, 2,
\infty\}\), and look at adversarial examples generated by projected gradient
descent as in \cite{madry2018towards}.
We also must specify (i) a dataset \(\cD \subset \cX \times \cY\) of images, and (ii) an image classifer \(f: \cX \to \cY\).

Having made these decisions, for a fixed dimension \(d\) and constraint
\(\epsilon\) and for each data point \((x, y) \in \cD\), we sample a
\(d\)-dimensional subspace \(V \subseteq \cX\) according to the specified
distribution on \(\Gr(d, \dim \cX)\), generate an adversarial perturbation
\(x' = x + \delta \in \cX\) such that \(\delta\) is constrained to $V$ and with \(d(x', x) \leq \epsilon\) using the algorithm
\(\mathcal{A}\), and record whether the attack was successful, that is:
\(\mathbf{1}(f(x') \neq y)\). To obtain a low-variance estimate of adversarial
success we average \(\mathbf{1}(f(x') \neq y)\) over the dataset \(\cD\) (or a reasonably large subsample thereof) and
sample a different subspace \(V \subseteq \cX\) for each datapoint \((x, y)\) to approximate 
\begin{equation}
   \label{eq:advsucc-as-expect}
   \success(d, \epsilon) = P(f(x') \neq y).
\end{equation}
It should be emphasized that we are computing statistics for random subspaces; as other works discussed in \cref{sec:rw} have shown, there are specific subspaces in which a higher adversarial success rate can likely be achieved.

\section{Experiments}
\label{sec:exper}

\begin{figure*}[tb]
    \centering
   \begin{subfigure}{0.48\linewidth}
        \centering
        \includegraphics[width=\linewidth]{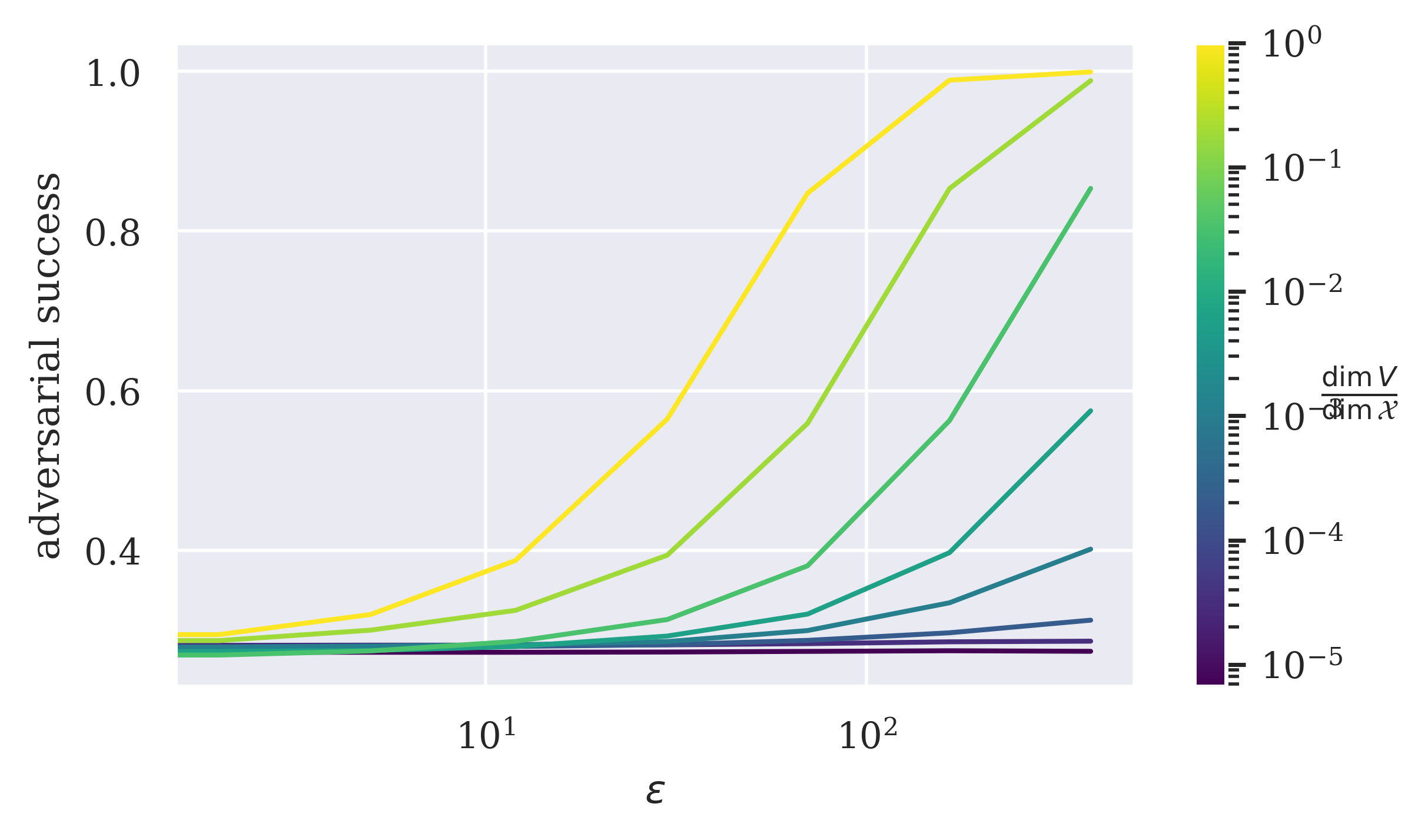}
    \end{subfigure}
    \begin{subfigure}{0.48\linewidth}
        \centering
        \includegraphics[width=\linewidth]{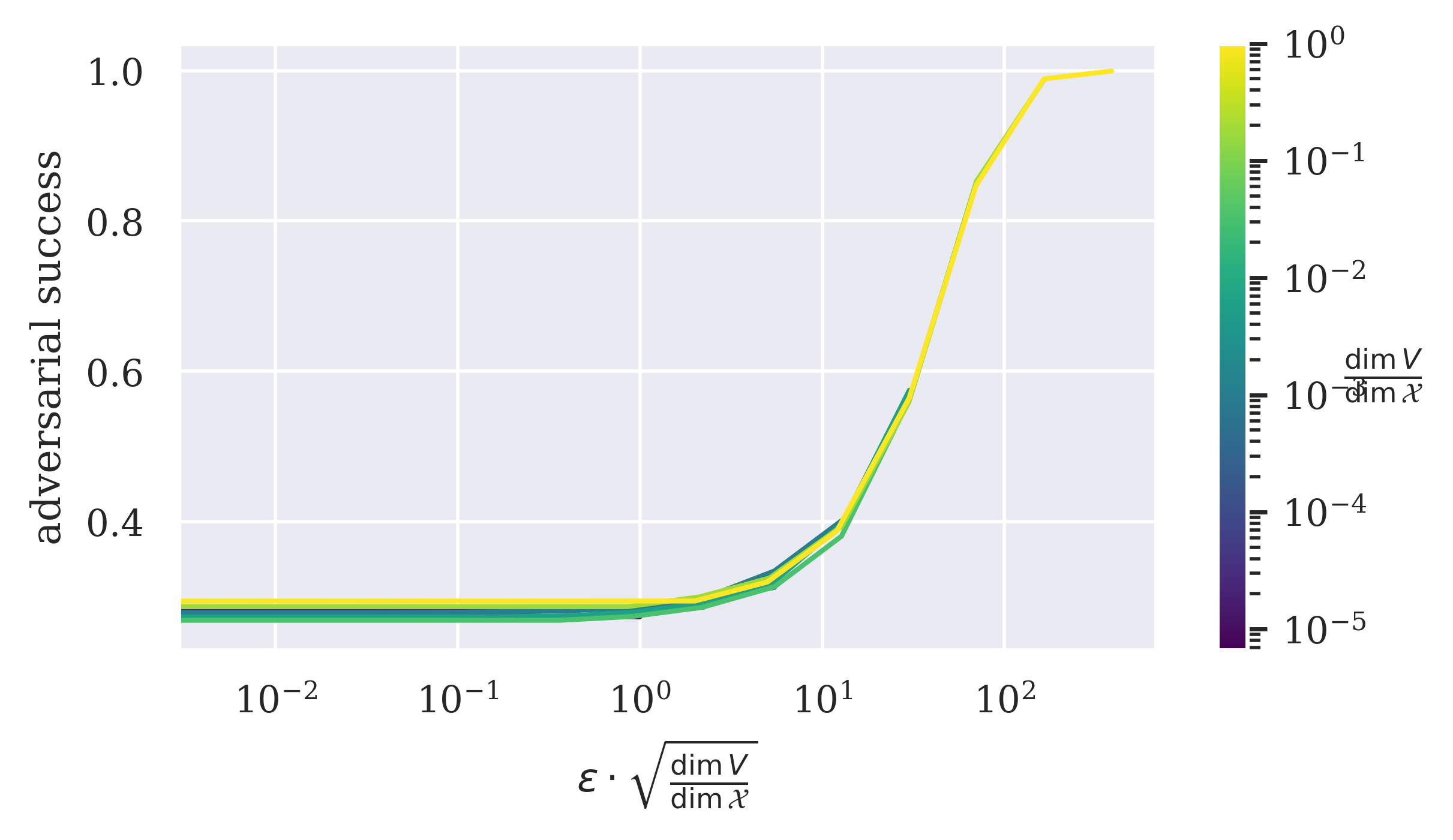}
    \end{subfigure}
    \caption{\textbf{Top}: success of PGD adversarial attacks on an ImageNet
    trained ResNet50, with $\ell^2$-norm constraints on perturbation budget (all
    other experimental details are as in \cref{fig:infty}, which displays
    analogous results for \(\ell^\infty\)-norm constraints).
    \textbf{Left}: these curves
    become aligned when we reparameterize the $x$-axis by scaling by
    $\sqrt{\frac{\dim V}{\dim \cX}}$.}
   \label{fig-L2-imagenet}
\end{figure*}

\subsection{Datasets and models}
We experiment with several image classification data sets and model
architectures, of increasing image resolution and network capacity:
\begin{itemize}
    \item The small convolutional network used in
    \cite{madry2018towards} trained on the MNIST dataset \cite{lecun1998mnist}.
    \item A ResNet9 \cite{He2016DeepRL} trained on the
    CIFAR10 dataset \cite{Krizhevsky2009LearningML}.
    \item A ResNet50 \cite{He2016DeepRL} trained on the
    ImageNet dataset \cite{imagenet_cvpr09}.
\end{itemize}
For further details on model architectures and training, we refer to \cref{sec:training}.

\subsection{Functional form of adversarial success}
\label{sec:eps-d-pinv}
We may view the adversarial successes \(\success(\dim V, \epsilon)\) as a sequence of
functions of \(\epsilon\), one for each \(\dim V \in \{1, \dots, \dim \cX\}\) as
shown in the top plot of \cref{fig:infty}, which displays results of \(\ell^\infty
\) PGD adversarial attacks
constrained to spans of subsets of standard basis vectors for a ResNet50 trained on ImageNet. It appears that for
varying \(\dim V\), the curves \(\success(\dim V, \epsilon)\) differ by \(x\)-axis
scalings, that is, transformations of the form \(\success(\dim V, \epsilon) \gets
\success(\dim V, \lambda \epsilon)\) for some \(\lambda > 0\). This is indeed the
case: the left plot in \cref{fig:infty} shows that the curves \(\success(\dim V, \epsilon \cdot
\frac{\dim V}{\dim \cX})\) are almost identical. \Cref{fig:infty-MNIST,fig:infty-CIFAR}
show the same phenomena for a 2-layer CNN on MNIST and a ResNet9 on CIFAR10. 

We can think of this analysis as expressing a \emph{decomposition} \(\success(\dim V, \epsilon) = g(\epsilon \cdot \frac{\dim V}{\dim \cX})\) into a composition of two functions, the first being the map \((\dim V, \epsilon) \mapsto \epsilon \cdot \frac{\dim V}{\dim \cX}\), the second map \(g\) being some single-variable function applied to \(\epsilon \cdot \frac{\dim V}{\dim \cX} \). We do not attempt to identify this \(g\).

\Cref{fig-L2-imagenet} shows results analogous to those of \cref{fig:infty} for PGD adversarial examples constrained in
the \(\ell^2\)-norm (with plots for other architecture types and models in \Cref{fig-L2-MNIST} and \Cref{fig-L2-CIFAR10}). In this case, the reparametrization of the \(x\)-axis that
results in almost identical curves is obtained by replacing \(\success( \dim V,
\epsilon)\) with \(\success(\dim V, \epsilon \cdot \sqrt{\frac{\dim V}{\dim \cX}})\). This shows that the
functional form of adversarial success in terms of \(\epsilon \) and \(\dim V\)
depends on the norm constraining adversarial perturbations. In the following
section, we argue that as long as \(p > 1\) adversarial success with \(\ell^p\)-constraints depends on \(\epsilon d^{\frac{1}{q}} \) where \(\frac{1}{p} +
\frac{1}{q} = 1\), a hypothesis consistent with experimental results in \cref{fig-L2-imagenet} and \cref{fig:infty}. 

The case \(p=1\) is more complicated, and we defer its analysis to \cref{sec:an1nrm}.

\section{Comparison with existing theoretical predictions}
\label{sec:comp-w-thry}

There are many existing works investigating the mathematical source of
adversarial examples for deep learning models. Several of these include (either as
a main result or a byproduct of calculations) predictions for the functional
form of adversarial success in terms of perturbation budget \(\epsilon\) and the
dimension \(\dim V\) to which perturbations are constrained. We reviewed a subsample
of such papers. Note that most of these papers (the exceptions being \cite{adv2random,GuoOrigins}) focus on the dimension of the input space of a model alone and do not consider the additional constraint that adversarial examples are confined to a subspace.
\begin{enumerate}[(i)]
    \item \label{item:1} From the analysis in \cite{harness} one can predict that adversarial success is a function of  \(\epsilon \frac{\dim V}{\dim X}\) (for \(p=\infty\)). 
    \item \label{item:2} \cite{adv2random,cubukIntriguingPropertiesAdversarial2017,GuoOrigins}
     predict that adversarial success is a function of \(\epsilon
    (\frac{\dim V}{\dim \cX})^{\frac{1}{2}}\) (for \(p=2\)).
    \item \label{item:3} From the analysis in \cite{MoosaviDezfooli2016DeepFoolAS,pmlr-v97-simon-gabriel19a} one can \emph{predict}
    that adversarial success is a function of \(\epsilon (\frac{\dim V}{\dim \cX})^{\frac{1}{q}}\) (for
    \(\frac{1}{p} + \frac{1}{q} = 1\)). However, arriving at such a functional
    form from the derivations in
    \cite{MoosaviDezfooli2016DeepFoolAS,pmlr-v97-simon-gabriel19a} requires
    multiple non-trivial steps not carried out in those works.
    \item \label{item:4} \cite{shafahi2018are} predicts that
    adversarial success is a function of \(\epsilon \dim V^{\frac{1}{2} -
    \frac{1}{\min(2, p)}}\). 
\end{enumerate}
The predictions of \cref{item:1,item:2,item:3} are all consistent with our
experimental results, suggesting that the situation where adversarial examples are constrained to a subspace of dimension $d$ is effectively equivalent to the unconstrained situation where data is found in an ambient space of dimension $d$. Those of \cref{item:4} are not obviously consistent with
our experimental data, although we refrain from saying that they are
inconsistent since the analysis of \cite{shafahi2018are} involves a series of
inequalities, and it is unclear how the predictions would change if one used
slightly different approximations.\footnote{Moreover, the aim of
\cite{shafahi2018are} was to demonstrate prevalence of adversarial examples, not
to estimate functional forms for adversarial success.}

The dependence of adversarial success on \(\epsilon (\frac{\dim V}{\dim \cX})^{\frac{1}{q}}\) can be
derived from the simplest possible toy model, namely binary linear
classification. Let \(\cX = \RR^n\) and suppose 
\begin{equation}
    f(x) = w^T x + b, \text{ for some  } w\in \RR^n, b \in \RR.
\end{equation}
Let  \(V \subseteq \RR^n \) be a subspace with \(\dim V = d\). We will assume
there exists an isometry \(U: \RR^d \xrightarrow{\simeq} V\) with respect to the
\(\ell^p \) metric.\footnote{This assumption holds in all of our experiments.}
The point \(x\) admits an \(\ell^p\) adversarial example in the subspace \(V \)
with budget \(\epsilon\), i.e. there is a \(\delta \in V\) such that \(\sign
f(x+\delta) \neq \sign f(x)\) and \(\nrm{\delta}_p \leq \epsilon\), if and only
if the \emph{\(\ell^p\) margin of \(x\)}
\begin{equation}
    \min_{\delta \in V} \big\{\nrm{\delta}_p \; \big| \;  f(x+ \delta) = 0\big\}
\end{equation}
is at most \(\epsilon\).
\begin{lemma}
\label{lem:margin}
    With the above definitions and notation and with \(\frac{1}{p} + \frac{1}{q}
    = 1\),
    \begin{equation}
        \min_{\delta \in V} \big\{\nrm{\delta}_p \; \big| \;  f(x+ \delta) = 0\big\} = \frac{\nrm{w^T x + b}}{\nrm{w^T U}_q}.
    \end{equation}
\end{lemma}
\Cref{sec:deriv} contains a proof. Our experimental results only measure the \emph{probability} that \(x\) admits
an \(\ell^p\)-adversarial example in the subspace \(V \) with budget
\(\epsilon\). By the above lemma, this probability is \(P(\frac{\nrm{w^T x +
b}}{\nrm{w^T U}_q} \leq \epsilon)\), which can be rewritten as \(P(\nrm{w^T x +
b} \leq \epsilon \nrm{w^T U}_q)\). We claim that when \(p>1\) and \(V \) (equivalently \(U \)) is sampled with
sufficient randomness
\begin{equation}
    \label{eq:claim}
    \mathbb{E}\big[\nrm{w^T U}_q\big] = \Big(\frac{d}{n}\Big)^{\frac{1}{q}} \nrm{w}_q.
\end{equation}
In the case where \(V\) is generated by a subset, say \(\{e_{i_1}, \dots,
e_{i_d}\}\) of basis vectors, this can be argued as follows:
\begin{equation}
\label{eq:claim2}
    \begin{split}
        \frac{\nrm{w^T U}_q^q}{\nrm{w}_q^q} &= \frac{\sum_{j=1}^d \nrm{w_{i_j}}^q}{\sum_{i=1}^n \nrm{w_{i}}^q} \\
        &= \frac{d}{n} \frac{\frac{1}{d}\sum_{j=1}^d \nrm{w_{i_j}}^q}{ \frac{1}{n}\sum_{i=1}^n \nrm{w_{i}}^q} \\
    \end{split}
\end{equation}
When the basis subset \(\{e_{i_1}, \dots,
e_{i_d}\}\) is sampled uniformly\footnote{For example, by taking \(i_1,\dots,
i_d = \sigma(1),\dots, \sigma(d)\) where \(  \sigma \) is a uniformly random
permutation of \(\{1,\dots, n\}\).} we claim that that the expectation of the term
\(\frac{1}{d}\sum_{j=1}^d \nrm{w_{i_j}}^q\) is exactly \(\frac{1}{n}\sum_{i=1}^n
\nrm{w_{i}}^q\) (at least when \(d=1\) this is immediate). Thus after averaging
over many random subspaces \(V\),  
\begin{equation}
\label{eq:claim3}
    \mathbb{E}\Big[\frac{\frac{1}{d}\sum_{j=1}^d \nrm{w_{i_j}}^q}{ \frac{1}{n}\sum_{i=1}^n \nrm{w_{i}}^q}\Big]=1, \text{  hence  } \mathbb{E}\Big[\frac{\nrm{w^T U}_q^q}{\nrm{w}_q^q}\Big] = \frac{d}{n}.
\end{equation} 
Taking \(q\)-th roots and rearranging gives \cref{eq:claim}. 

Note that in our experiments we compute something analogous to \(P(\nrm{w^T x +
b} \leq \epsilon \nrm{w^T U}_q)\) where probability is with respect to the
underlying distribution of \(x \) and choice of \(U \). Using a ``point
estimate'' and replacing \(\nrm{w^T U}_q\) with its mean
\((\frac{d}{n})^{\frac{1}{q}} \nrm{w}_q\), one would simplify to \(P(\nrm{w^T x +
b} \leq \epsilon (\frac{d}{n})^{\frac{1}{q}} \nrm{w}_q)\), which since we treat
\(w, b\) and the distribution of \(x\) as given is a function of \(\epsilon (\frac{d}{n})^{\frac{1}{q}}\).

When \(p = 1\), so \( q = \infty\), \cref{lem:margin} remains valid but the tricks applied in \cref{eq:claim2,eq:claim3} do not make sense, and indeed our experimental results in \cref{sec:an1nrm} suggest dependence of adversarial success on \(\epsilon (\frac{\dim V}{\dim \cX})^{\frac{1}{q}}\) alone breaks down somewhat in this case. For further analysis of this case, we refer to \cref{sec:an1nrm}. 

\section{Limitations}

Adversarial examples given by gradient-based perturbations with \(\ell^p\)
constraints make up only one (and arguably, a narrow) type of
distribution-shifted test data causing machine learning model failure. For
further discussion of this point see
\cite{gilmerMotivatingRulesGame2018a,gilmerDiscussionAdversarialExamples2019a}.
While we take inspiration from adversarial example generators
constraining perturbations to subspaces (surveyed in \cref{sec:rw}), our
experiments are limited to the baseline of random subspace selection (whereas
most subspace-constrained adversarial example generators choose their subspace
more carefully). We also only experiment with image classifiers, though adversarial
examples have been found to exist for essentially all deep learning systems \cite{maus2023adversarial,kuppa2019black,xie2017adversarial}. 

\section{Conclusion and open questions}

We demonstrate that the adversarial success \(\success(\dim V, \epsilon)\) of PGD
attacks constrained to a (random) \(\dim V\)-dimensional subspace \(V \) of the model input space \(\cX \) with \(\ell^p\)
budget \(\epsilon\) (and \(p>1\)) is essentially a function of the single variable  \(\epsilon (\frac{\dim V}{\dim \cX})^{\frac{1}{q}}\) where \(\frac{1}{p} + \frac{1}{q} = 1\) (rather than a
function of two variables as considered in prior work). The fact that this
relationship can be derived in the toy example of a linear binary classifier,
and holds quite sharply in all our experiments, seems to lend further credence
to the theory that adversarial examples are a byproduct of the locally linear
behavior of neural networks with high dimensional input spaces.


\subsubsection*{Acknowledgments}
The research described in this paper was conducted under the Laboratory Directed
Research and Development Program at Pacific Northwest National Laboratory, a
multiprogram national laboratory operated by Battelle for the U.S. Department of
Energy.

%% file: appendix.tex
\section{Derivations}
\label{sec:deriv}

\begin{proof}[Proof of \cref{lem:margin}]
Recall that our goal is to solve the constrained optimization problem
\begin{equation}
\label{eq:min-prob}
    \min\{\nrm{\delta}_p \, | \, \delta \in \RR^d, w^T (x + U \delta) + b=0\}
\end{equation}
Using the method of Lagrange multipliers, we know that a minimizer \(\delta \in \RR^d \) must satisfy the critical point condition
\begin{equation}
\label{eq:criticalpt}
    \lambda U^T w \in \partial \nrm{\delta}_p 
\end{equation}
where \( \partial \nrm{\delta}_p  \) is the subdifferential of the \(\ell^p \)-norm at \(\delta \in \RR^d \). It is a classical fact (see e.g. \cite[Prop. 1.2]{bachOptimizationSparsityInducingPenalties2011}) that
\begin{equation}
\label{eq:subdiff}
    \partial \nrm{\delta}_p = \begin{cases}
    \{v \in \RR^d \, |\, \nrm{v}_q \leq 1 \} & \text{if } \delta = 0 \\
    \{v \in \RR^d \, |\, \nrm{v}_q = 1  \text{ and } v^T \delta = \nrm{\delta}_p \} & \text{otherwise.}
    \end{cases}
\end{equation}
If the first case occurs, clearly the minimum of \cref{eq:min-prob} is \(0 \), and since \(\delta = 0 \) we obtain 
\begin{equation}
    0 = w^T (x + U \delta) + b = w^T x + b,
\end{equation}
hence in this case \(\frac{\nrm{w^T x + b}}{\nrm{w^T U}_q} = 0\) as well and the lemma holds. 

In the case where \(\delta \neq 0 \), combining \cref{eq:criticalpt,eq:subdiff} we see that for some \(\lambda \)
\begin{equation}
    \nrm{\lambda U^T w}_q = 1 \text{  and  } \lambda w^T U \delta = \nrm{\delta}_p;
\end{equation}
from the first equation we immediately identify \(\nrm{\lambda} = \nrm{U^T w}_q^{-1} \), and taking absolute values on both sides of the second then gives 
\begin{equation}
    \begin{split}
        \nrm{\delta}_p &= \nrm{\lambda} \nrm{w^T U \delta} \\
        &= \frac{ \nrm{w^T U \delta}}{\nrm{U^T w}_q}.
    \end{split}
\end{equation}
Finally, recalling \(w^T (x + U \delta) + b=0\) gives \(\nrm{w^T U \delta} = \nrm{w^T x + b} \), completing the proof.
\end{proof}

\section{Experimental details}
\label{sec:exp-det}

\subsection{Model architectures and training details}
\label{sec:training}

For our MNIST experiments we use the simple 2 layer convolutional network of
\cite{madry2018towards} --- we ported the TensorFlow code available at
\href{https://github.com/MadryLab/mnist_challenge}{https://github.com/MadryLab/mnist\_challenge}
to PyTorch \cite{torch}. We train it using SGD with momentum 0.9, batch size
\(1024\) and weight
decay \(10^{-4}\) for 100 epochs, with initial learning rate \(10^{-3}\) and
learning rate drops by a factor of \(0.1\) whenever validation accuracy doesn't
improve by \(1\%\) for 10 epochs. We save the weights with the best validation
accuracy (\(\approx 98.95 \%\)).

For our CIFAR10 experiments we use the ResNet9 from MosaicML's Composer library
\cite{mosaicml2022composer}. We train it with SGD with momentum 0.9, batch size
\(512\) and weight decay \(10^{-4}\) for 160 epochs, with initial learning rate \(10^{-1}\) and
learning rate drops by a factor of \(0.1\) whenever validation accuracy doesn't
improve by \(1\%\) for 10 epochs. We save the weights with the best validation
accuracy (\(\approx 91.72 \%\)). 

For our ImageNet experiments we use the ResNet50 from TorchVision
\cite{torchvision}. We train it with SGD with momentum 0.9 and weight decay
\(10^{-4}\) for 100 epochs, with initial learning rate \(1.0\) and learning rate
drops by a factor of \(0.1\) whenever validation accuracy doesn't improve by
\(1\%\) for 10 epochs. Due to distributed data parallel training with batches of
size 512 on each of 8 GPUs, our effective batch size is \(8 \cdot 512 = 4096\).
We save the weights with the best validation accuracy (\(\approx 72.84 \%\)). 

\subsection{Tuning PGD step sizes}

In our experiments we generate a large number of PGD adversarial examples for a
wide range of perturbation constraints \(\epsilon\) and in subspaces of varying
dimension. In order for our numerical experiments to address our questions about
the behavior of \( \success(d, \epsilon) \), it is \emph{crucial} that our PGD
algorithm for optimizing \(\delta\) has the capacity to achieve the boundary
case \(\nrm{\delta}_p = \epsilon\). We found that with some standard choices of
step size, this did not occur, resulting in an unpleasant situation where the
effective budget was significantly lower than \(\epsilon\) simply due to a
too-small PGD step size. Here we briefly discuss a principled choice of PGD step
size that accounts for the dimension \(d\) of the subspace to which \(\delta\)
is constrained. First we must specify the PGD algorithm being used.

Our basic PGD implementation (adapted from \cite{madry2018towards}) iterates the
following: we constrain \(\delta \) to a \(d\)-dimensional subspace \(V
\subseteq \cX \) using an isometry \(U: \RR^d \to V \) as in
\cref{sec:comp-w-thry}, and  initialize \(\delta_0 = 0\) 
. 
Then, for \(t = 1, \dots, T\) where \(T \) is the maximum number of
steps, we let \(g_t = \nabla_{\delta} \ell(f(x+ U \delta), y)\) where \(\ell \) is
cross entropy, and replace it with the ``normalized'' gradient
\begin{equation}
\label{eq:fgsm}
    \tilde{g}_t := \begin{cases}
        \frac{g_t}{\nrm{g_t}_p}, & p\in \{1,2\} \\
        \sign{g_t}, & p=\infty.
    \end{cases}
\end{equation}
We then project \(\delta_{t-1} + \eta \tilde{g}_t \), where \(\eta\) is a
learning rate, onto the \(\ell^p\) \(\epsilon\)-ball centered at \(0\) to obtain
in the case \(p\in \{1,2\}\)
\begin{equation}
    \delta_{t} = \begin{cases}
        \epsilon \frac{\delta_{t-1} + \eta \tilde{g}_t}{\nrm{\delta_{t-1} + \eta \tilde{g}_t}_p}, & \nrm{\delta_{t-1} + \eta \tilde{g}_t}_p > \epsilon \\
        \delta_{t-1} + \eta \tilde{g}_t, & \text{otherwise}
    \end{cases}
\end{equation}
and in the case \(p=\infty\)
\begin{equation}
    \delta_{t} = \begin{cases}
        \epsilon \frac{\delta_{t-1} + \eta \tilde{g}_t}{\max\{\delta_{t-1} + \eta \tilde{g}_t\}}, & \max\{\delta_{t-1} + \eta \tilde{g}_t\} > \epsilon \\
        \delta_{t-1} + \eta \tilde{g}_t, & \text{otherwise}.
    \end{cases}
\end{equation}
Finally, we must ensure that \(x + U\delta\) lies in the image hypercube \([0,
1]^{C \times H \times W}\) (in our implementation pixel values lie in \([0,
1]\)). To do this, we let \(\clip: \RR \to [0,1]\) be the clipping
function, i.e. \(\clip(x) = \max\{0, \min\{x, 1\}\}\), and replace \(\delta
\) with 
\begin{equation}
    U^T (\clip(x + U \delta) - x)
\end{equation}
(here we use the fact that \(U\) is orthogonal and so \(U^T\) is a left inverse
for \(U\)).

To set the step size \(\eta\), we can adopt the heuristic that the \(\delta_t\)
behave like a random walk, i.e. that the normalized gradients \(\tilde{g}_t \)
are sampled IID from some distribution (this almost certainly quite false, but
we found it to be useful in a back-of-the-envelope sort of way). We will even
further assume  that for each \(t \) the coordinates of \(\tilde{g}_t \) are
IID. Ignoring projection and clipping, we have \(\delta_T = \eta \sum_{t=1}^T
\tilde{g}_t\). In the case \(p=2\), using the supposed IID-ness we see that 
\begin{equation}
    E[\nrm{\delta}_2^2] = \eta^2 \sum_{t=1}^T \nrm{\tilde{g}_t}_2^2 = \eta^2 T.
\end{equation}
Since we want to ensure the left hand side is at least \(\epsilon\), we obtain
the step size
\begin{equation}
    \eta = \frac{\epsilon}{\sqrt{T}}.
\end{equation}
In practice we multiply the above by 2 for good measure.\footnote{Note
that this is \emph{larger} than what is suggested in \cite[p. 12, section
``Resistance for different values ...'']{madry2018towards}, which divides by
\(T\). By ``for good measure'' we mean that our primary concern is using \emph{too small} of a step size.} In the case \(p=\infty\),  by our IID-ness assumptions and the fact
that by definition \(\tilde{g}_t = \sign{g_t}\), the individual coordinates
\(\tilde{g}_{t, j}\) for \(j = 1,\dots, d\) are IID samples from \(\{\pm 1\}\)
and so 
\begin{equation}
    \label{eq:max-thing}
    \nrm{\delta_T}_{\infty} =\eta \sqrt{T} \max_{j=1,\dots, d} \{\nrm{\frac{1}{\sqrt{T}}\sum_{t=1}^T \tilde{g}_{t, j}}\}
\end{equation}
The distribution of each term \(\frac{1}{\sqrt{T}}\sum_{t=1}^T \tilde{g}_{t,
j}\) tends towards a Gaussian distribution with mean 0 and variance \(1\) as \(T
\to \infty \) by the
central limit theorem.
Letting \(\Phi(x)\) be the standard normal CDF, the CDF of each
 \(\nrm{\frac{1}{\sqrt{T}}\sum_{t=1}^T \tilde{g}_{t, j}}\) is approximated by
\begin{equation}
    \label{eq:cdf}
    F(x):=\Phi(x) - \Phi(-x).
\end{equation}
By \cite{bahadurNoteQuantilesLarge1966}, the
distribution of the max occuring in \cref{eq:max-thing} is concentrated around
the \(\frac{d-1}{d}\)-th quantile of the CDF \cref{eq:cdf}, i.e.
\(F^{-1}(1-\frac{1}{d})\). Assuming \(d \) is relatively large, so that
\(1-\frac{1}{d}\) is near \(1\), we ignore the \(\Phi(-x)\) term in
\cref{eq:cdf} for the purposes of inversion and get
\begin{equation}
    F^{-1}(1-\frac{1}{d}) \approx \Phi^{-1}(1-\frac{1}{d})
\end{equation}

Recall that our objective is to ensure that
\(\nrm{\delta_T}_{\infty} \geq \epsilon\). By the above arguments, this
translates to 
\begin{equation}
    \label{eq:linf-step-size}
    \begin{split}
        \epsilon &\leq \eta \sqrt{T} \Phi^{-1}(1-\frac{1}{d}),  \text{  i.e. }\\
        \eta &\approx \frac{\epsilon}{\sqrt{T} \Phi^{-1}(1-\frac{1}{d})}.
    \end{split}
\end{equation}
Again, in practice we multiply by 2 for good measure. Observe that while our
\(\ell^2\) step size is independent of \(d\), \cref{eq:linf-step-size} does
depend on \(d\). In fact, as \(d \to \infty\) the step size goes to \(0\), but
very slowly (for \(d = 3\cdot 224^2\), the dimension of ImageNet images,
\(\Phi^{-1}(1-\frac{1}{d}) \approx 4.36\)). 

For the \(p=1\) case, we use a heuristic similar to that of \(p=2\) above; explicitly, we set 
\[ \eta = \sqrt{2\pi } \frac{\epsilon}{\sqrt{T}}  \]

\subsection{Adversarial example generation}

For each dataset and model, we select a range of perturbation budgets
\(\epsilon\) and subspace dimensions \(d\), in both cases logarithmically spaced
between minimum and maximal values of \(\epsilon\) and \(d\), with as many
grid points as we can afford (for MNIST and CIFAR, 32 different values of each,
for ImageNet only 8 of each). 

For each pair \((\epsilon, d)\) we loop over the entire validation set of the
relevant dataset, with the exception of ImageNet where we randomly sample 10,000
of the 50,000 images. We randomly sample a distinct subspace \(V_i \subset \cX\)
for each validation datapoint \((x_i, y_i) \) (as above, by randomly generating
a matrix \(U\) whose columns span \(V\)). We then loop through validation
datapoints  \((x_i, y_i) \) and corresponding matrices \(U_i\) and compute PGD
adversarial examples as described above, with \(T=16\) steps. We compute the
error over the  validation set (subsampled in the case of ImageNet), i.e. 
\[ 1 - \frac{1}{N} \sum_{i=1}^N \mathbf{1}(f(x_i + U\delta_i)=y_i). \]


\subsection{Subspaces sampled uniformly from the Grassmannian}

\begin{figure*}[tb]
    \centering
   \begin{subfigure}{0.45\linewidth}
      \centering
      \includegraphics[width=\linewidth]{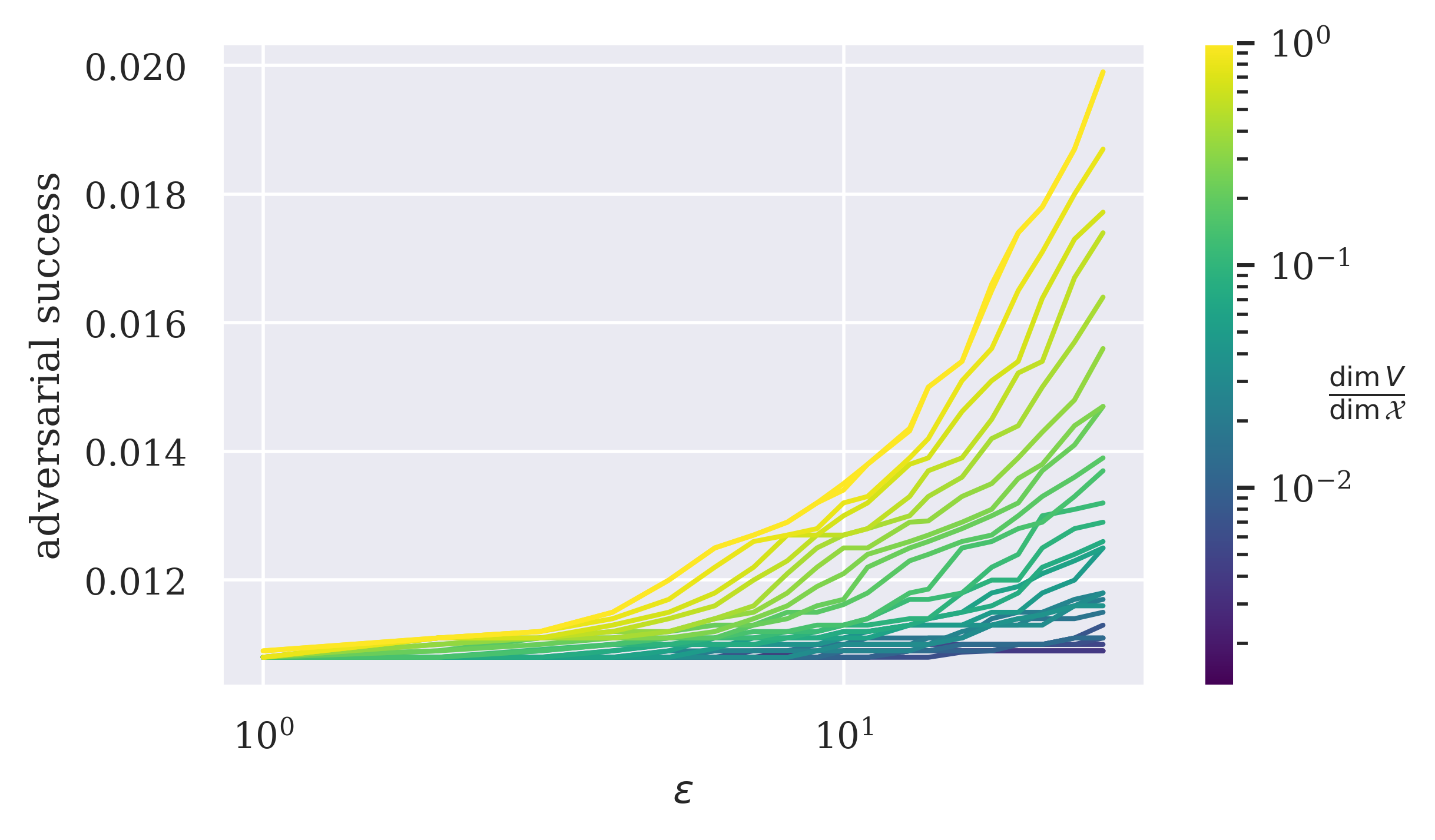}
      \caption[]{}\label{fig:mn2-a}
   \end{subfigure}
   \begin{subfigure}{0.45\linewidth}
      \centering
      \includegraphics[width=\linewidth]{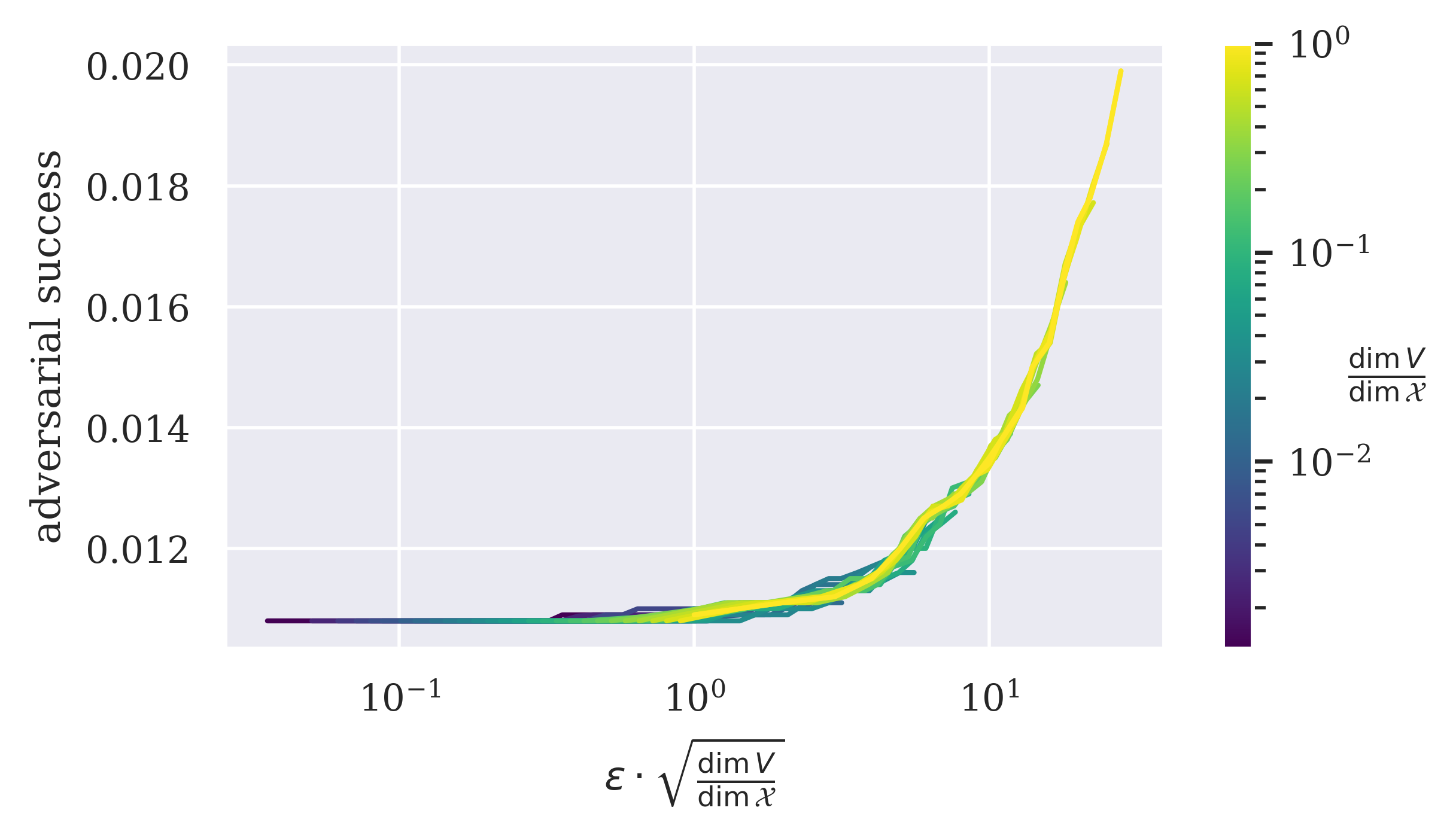}
      \caption[]{}\label{fig:mn2-b}
   \end{subfigure}
   \caption[]{\textbf{(a)} Projected gradient descent (PGD)
   adversarial examples for the 2-layer convolutional neural network of
   \cite{madry2018towards} on the MNIST validation set. In these experiments we
   generated the subspaces \(V\) by sampling uniformly from the Grassmannian, and
   constrain perturbations using the \(\ell^2 \) norm. \textbf{(b)}
   The same data,  reparametrized by plotting \(\epsilon \cdot \sqrt{\frac{\dim
   V}{\dim \cX}}\) along the \(x\)-axis. \emph{Note}:
   \(x\)-axes are log-scale.}
   \label{fig:mn2}
\end{figure*}

\begin{figure*}[htb]
    \centering
   \begin{subfigure}{0.45\linewidth}
      \centering
      \includegraphics[width=\linewidth]{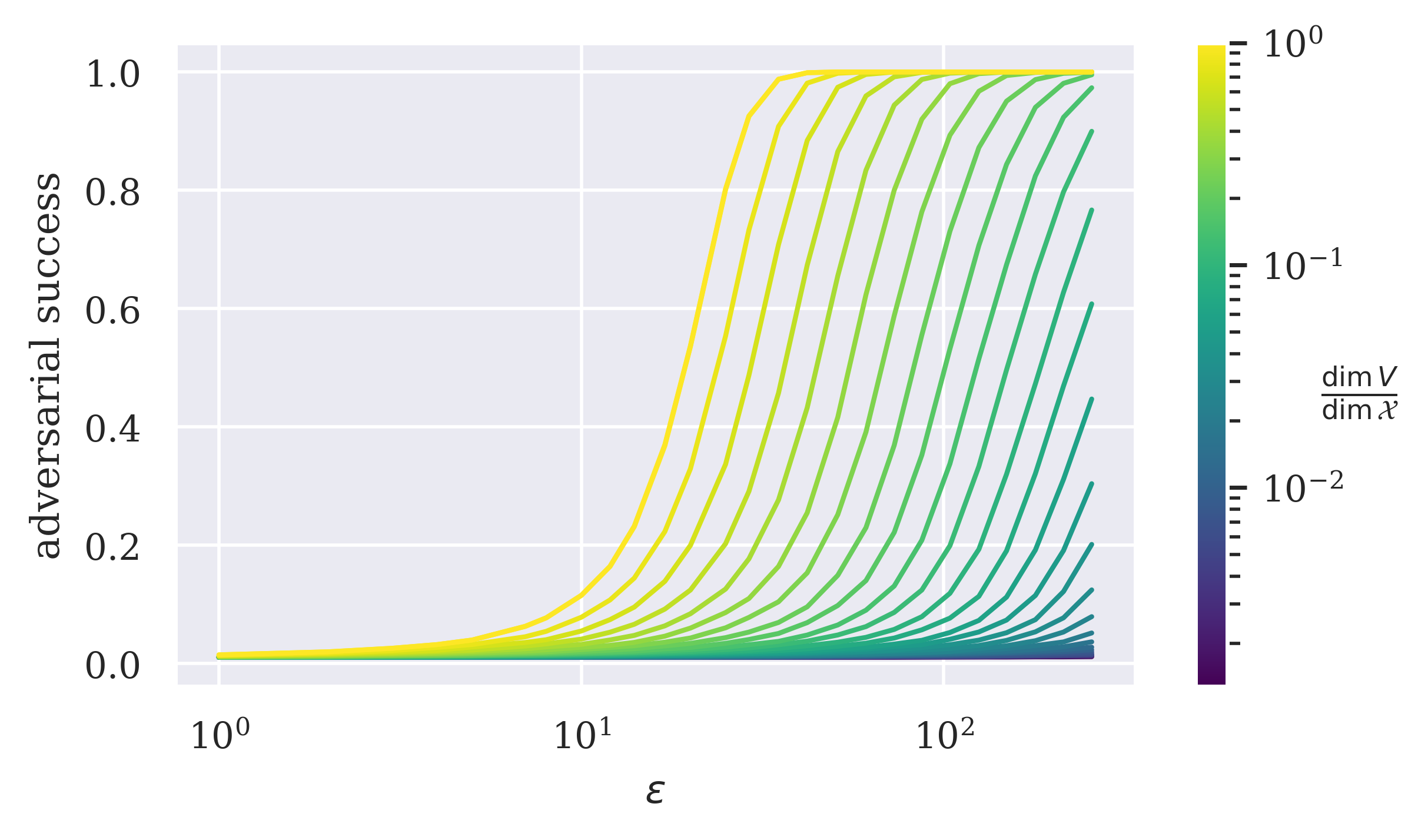}
      \caption[]{}\label{fig:infty-a}
   \end{subfigure}
   \begin{subfigure}{0.45\linewidth}
      \centering
      \includegraphics[width=\linewidth]{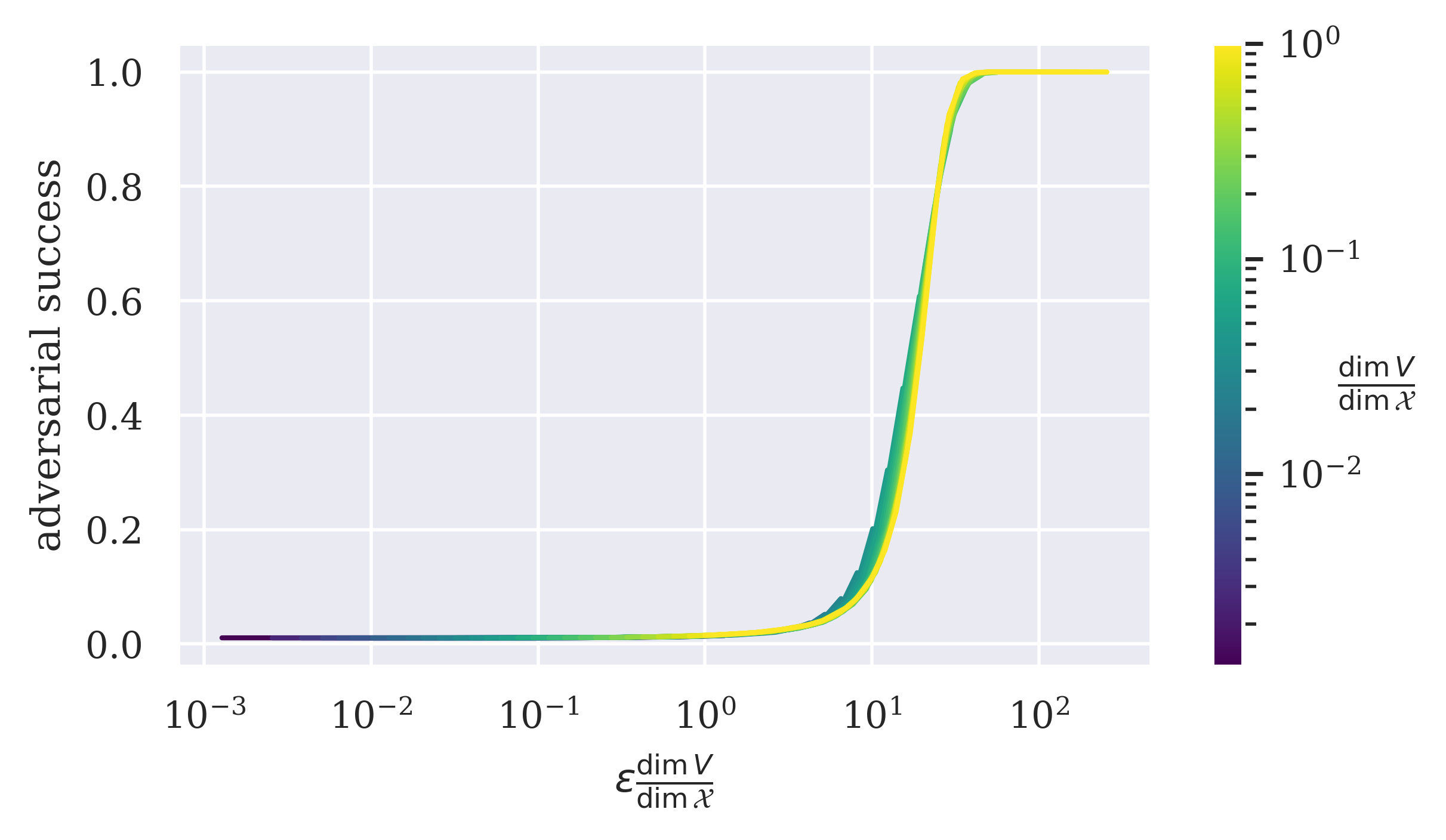}
      \caption[]{}\label{fig:infty-b}
   \end{subfigure}
   \caption{Plot for experiments analogous to those found in Figure \ref{fig:infty} but run with a 2-layer CNN trained and evaluated on MNIST.}
   \label{fig:infty-MNIST}
\end{figure*}

In the case of MNIST and for \(p=2\), we can also sample subspaces uniformly from the
Grassmannian \(\Gr(d, \dim \cX)\) by sampling matrices \(U\) of shape \(n\times
d\) with orthonormal columns using the QR decomposition as in used in the method
\texttt{scipy.stats.ortho\_group} of \cite{2020SciPy-NMeth}. The results, shown
in \cref{fig:mn2}, are similar to those in \cref{fig:infty-a,fig:infty-b}. 

\subsection{Analysis of the 1-norm case}
\label{sec:an1nrm}
\cwg{TODO: new 1 norm plots, modify analysis as necessary}
When \(p = 1 \), \( q = \infty \) and so the arguments used in \cref{eq:claim2,eq:claim3} do not make sense as written; moreover, while we have not explicitly verified this, it seems that attempting to take a limit of those equations as \(q \to \infty \) one will encounter an ``\(\infty/\infty\)'' case, and it's not clear that e.g. L'Hospital's rule helps at all.

Instead, we propose a different estimate of the quotient 
\begin{equation}
    \frac{\nrm{w^T U}_\infty}{\nrm{w}_\infty}, 
\end{equation}
proceeding as follows. We will again assume, as is the case in our experiments, that \(U \) is obtained by subsampling basis vectors, say \(\{e_{i_1}, \dots, e_{i_d}\}\). Then 
\begin{equation}
    \frac{\nrm{w^T U}_\infty}{\nrm{w}_\infty} = \frac{\max_j \{ \nrm{w_{i_j}} \}}{\max_i \{ \nrm{w_{i}} \}}
\end{equation}
The question, then, is how much smaller the \( \max \) over a random \(\dim V \)-element subset of the absolute values \(\nrm{w_i}\) is than the \( \max \) over all \( \dim \cX \) of them. The need to make some assumption on the distribution the \(\nrm{w_i}\) are drawn from seems unavoidable at this point: we suppose the coefficients \(w_i \) come from a standard normal distribution, so that their absolute values come from a ``half-normal'' (equivalently \( \chi_1\)) distribution: if \(\Phi \) is the standard normal cumulative distribution function, with this assumption the  cumulative distribution function of the \(\nrm{w_i}\) is
\begin{equation}
    F(x):= \Phi(x) - \Phi(-x).
\end{equation}
We make a further crude estimate that the numerator and denominator are \(\max\)s of \emph{independent} samples of size \(\dim V \) and \(\dim \cX \) respectively;\footnote{This is of course quite false, as the numerator differs from the denominator by taking the max over a subsample. How can one deal with this step more realistically?} then the theory of quantiles in large samples \cite{bahadurNoteQuantilesLarge1966} suggests the estimates 
\begin{equation}
\begin{split}
    \max_j \{ \nrm{w_{i_j}} \} &\approx F^{-1}(1 - \frac{1}{\dim V}) \text{  and} \\
    \max_i \{ \nrm{w_{i}} \} &\approx F^{-1}(1 - \frac{1}{\dim \cX}) 
\end{split}
\end{equation}
leading to the overall estimate 
\begin{equation}
\label{eq:best-guess-l1}
    \frac{\nrm{w^T U}_\infty}{\nrm{w}_\infty} \approx \frac{F^{-1}(1 - \frac{1}{\dim V})}{F^{-1}(1 - \frac{1}{\dim \cX})}
\end{equation}
\begin{figure*}[tb]
    \centering
   \begin{subfigure}{0.45\linewidth}
        \centering
        \includegraphics[width=\linewidth]{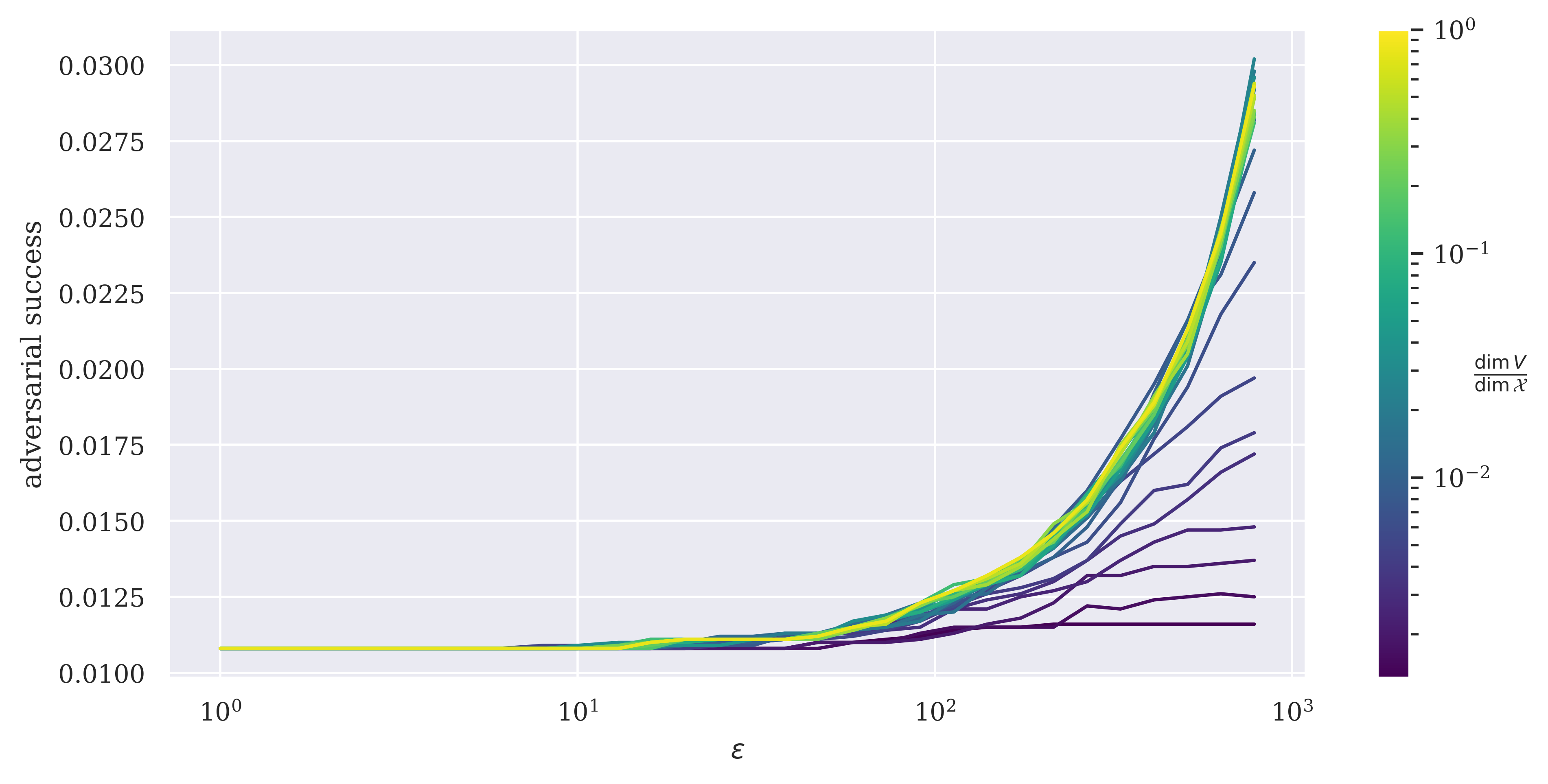}
        \caption[]{}\label{fig:mnist-1-a}
    \end{subfigure}
    \begin{subfigure}{0.45\linewidth}
        \centering
        \includegraphics[width=\linewidth]{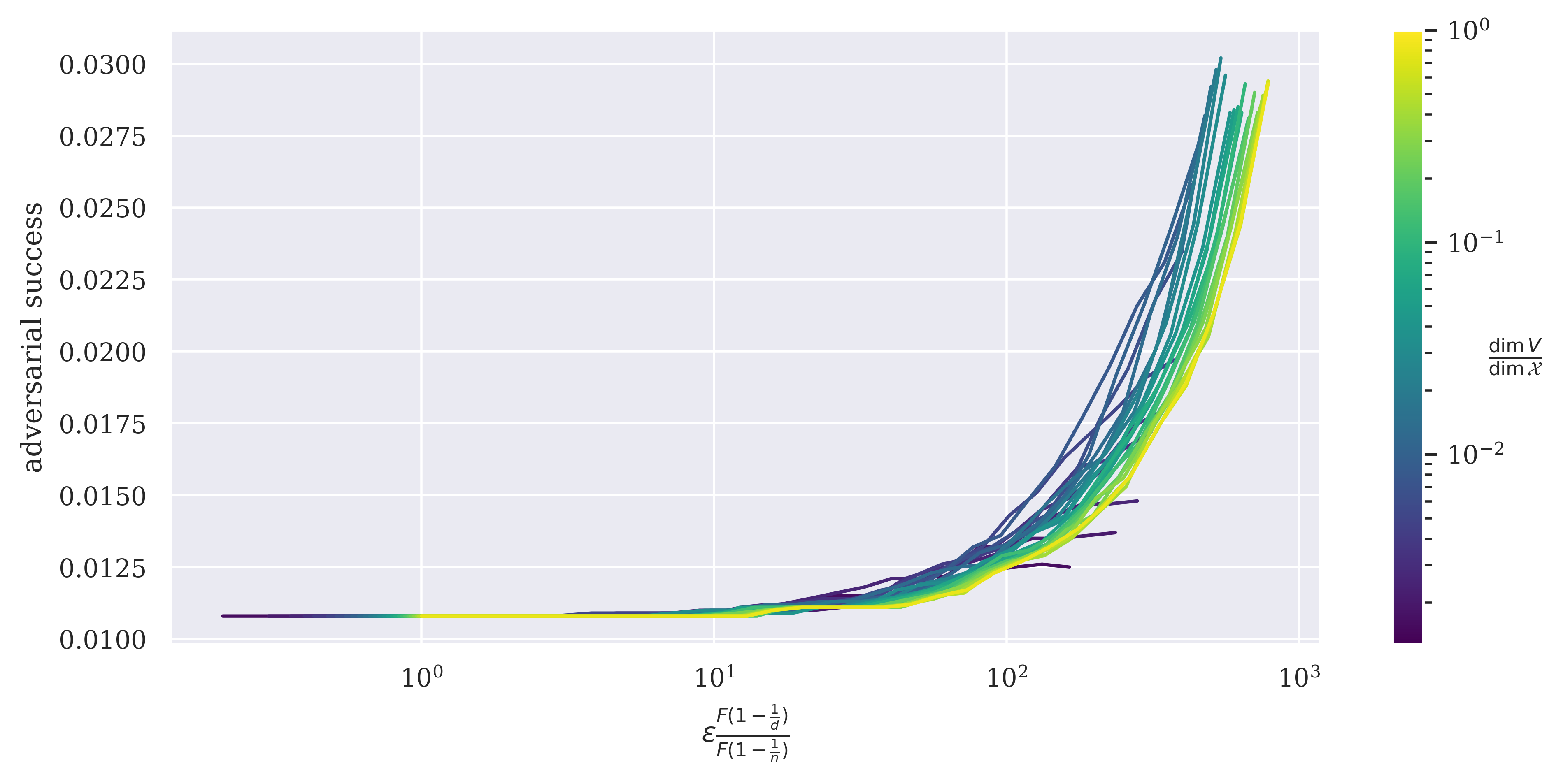}
        \caption[]{}\label{fig:mnist-1-b}
    \end{subfigure}
    \caption{\textbf{(a)} Success of PGD adversarial attacks on an MNIST trained small convolutional network, with $\ell^1$-norm constraints on perturbation budget, constrained to subspaces \(V \subseteq \cX\) spanned by \(\dim V\) randomly selected standard basis vectors. Adversarial examples are computed for all datapoints in the MNIST validation set. The $x$-axis is the $\epsilon$-bound used during example generation and the different colored curves indicate the dimension \(\dim V\) of the subspace to which the examples were constrained to, relative to the dimension \(\dim \cX \) (\(=28^2\)) of the ambient space. \textbf{(b)} These curves become \emph{more} aligned when we reparameterize the $x$-axis by scaling by $\frac{F(1-\frac{1}{\dim V})}{F(1-\frac{1}{\dim \cX})}$, where \(F\) is the cummulative distribution function of the absolute value of a standard normal random variable.}
   \label{fig-L1-MNIST}
\end{figure*}
\Cref{fig-L1-MNIST} shows the result of using \cref{eq:best-guess-l1} as a stand-in for \( (\frac{\dim V}{\dim \cX})^{\frac{1}{q}} \) in the case where \(p=1 \), on the MNIST dataset. One immediate observation is that at least for \(\dim V \) a significant fraction of \(\dim \cX \) (e.g. \(\frac{\dim V}{\dim \cX} \geq 0.1 \)) it does appear that the curves \( \success(\dim V, \epsilon ) \) converge to a common limit, as one would expect from a na\"ive application of a factorization \(\success(\dim V, \epsilon ) = g(\epsilon (\frac{\dim V}{\dim \cX})^{\frac{1}{\infty}}) = g(\epsilon) \). The reparameterization of \cref{eq:best-guess-l1} seems to do okay at accounting for behavior in the lowest dimensions, at the expense of over-compensating and pushing the curves corresponding to low-to-medium values of \(\dim V \) to the left of the curve corresponding to \(\dim V =\dim \cX \). There are various potential causes of this undesirable effect (roughly one per crude oversimplification in the above analysis). Results for CIFAR10 and ImageNet can be found in \cref{fig-L1-cifar} and \cref{fig-L1-imagenet} respectively. One concerning aspect of those two results is we see downturns in the curves \(\success(\dim V, \epsilon )  \) for the highest values of \(\epsilon \), suggesting there may have been issues with our PGD optimizer in the \(p=1\) case.  

One question we had was whether these results were impacted by sub-optimal PGD optimization. A reason for asking this is that the \(p=1\) case of \cref{eq:fgsm} is arguably incorrect: the ``correct'' way of deriving these generalized Fast Gradient Sign Method (FGSM) steps is through the analysis of \cref{sec:deriv}. Assuming \(U = I\) for simplicity, one sees that \(w^T \delta = \nrm{w}_\infty \nrm{\delta}_1 \), and one can show this occurs if and only if:
\begin{itemize}
    \item letting \(A = \mathrm{arg} \max_i \{\nrm{w_i}\} \subseteq \{1,\dots, n\} \) (the argmax of \(\nrm{w_i}\), which is a set in general although a single index with probability 1), \(\delta_i = 0 \) if \(i \notin A \).
    \item \(\sign \delta_i = \sign w_i \) for all \(i \in A \).
\end{itemize}
In the case where \(A = \{a\} \) (i.e. \(\nrm{w_i}\) has a unique maximum) we obtain the simplified solution \(\delta = (c \sign w_i) e_a \) (where \(e_a \) is the \(a\)-th standard basis vector. Hence for \(p=1 \), one can argue that we should use 
\begin{equation}
\label{eq:fgsm-1}
    \tilde{g}_t := e_{\mathrm{arg} \max_i \{\nrm{w_i}\}}
\end{equation}
We found that while this method performed similarly to that of \cref{eq:fgsm} for small values of \(\epsilon \), it struggled for large \(\epsilon \) and failed at the scale of ImageNet input space (see). A reasonable suspicion is that the number of basis directions selected by \cref{eq:fgsm-1} is bounded by the number of PGD iterations, and that when this number of iterations is far is smaller than the input dimension \cref{eq:fgsm-1} underexplores. However we leave further analysis to future work.

\section{Additional experimental results}
\label{sec:add-exper}

\begin{figure*}[htb]
    \centering
   \begin{subfigure}{0.45\linewidth}
      \centering
      \includegraphics[width=\linewidth]{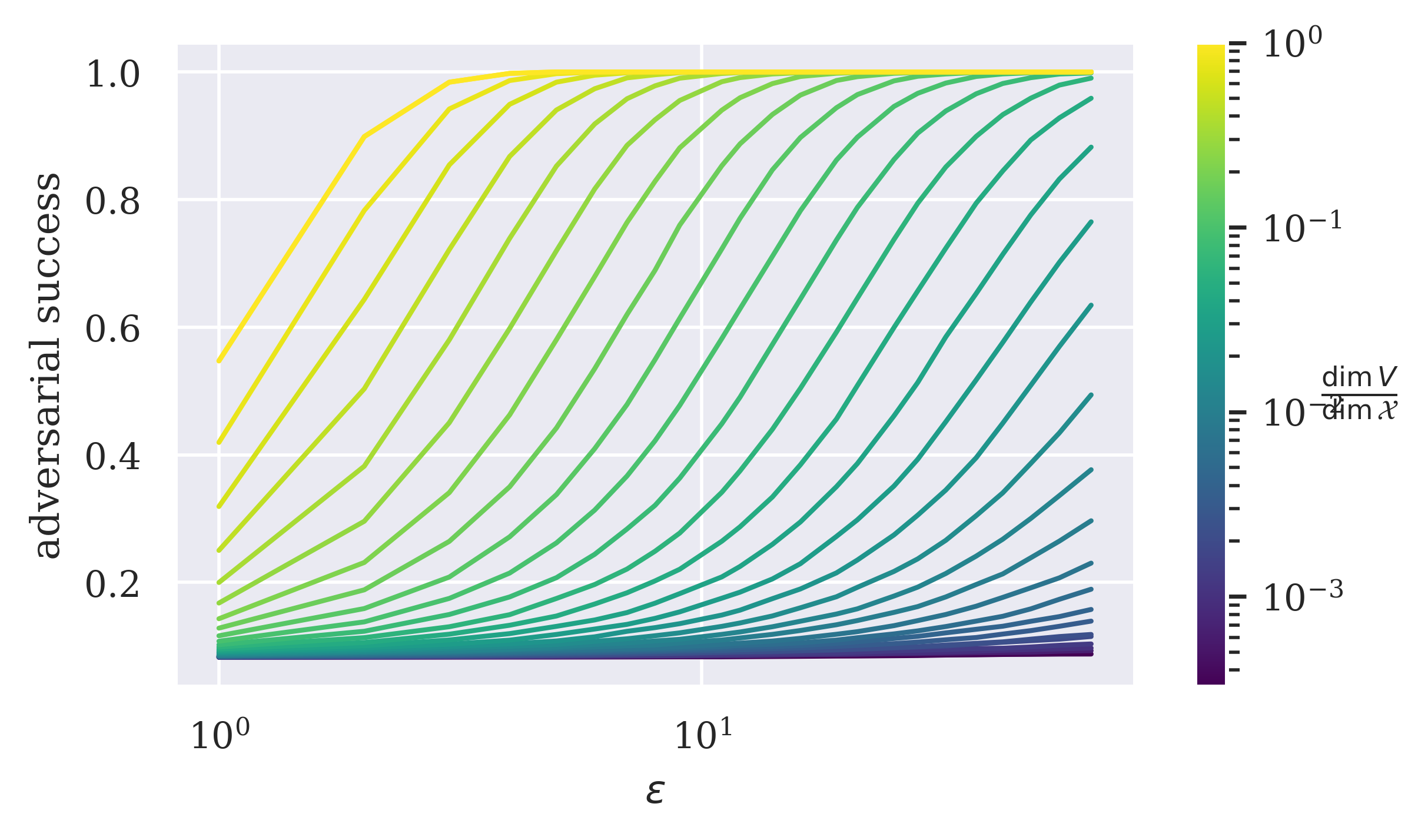}
      \caption[]{}\label{fig:infty-c}
   \end{subfigure}
   \begin{subfigure}{0.45\linewidth}
      \centering
      \includegraphics[width=\linewidth]{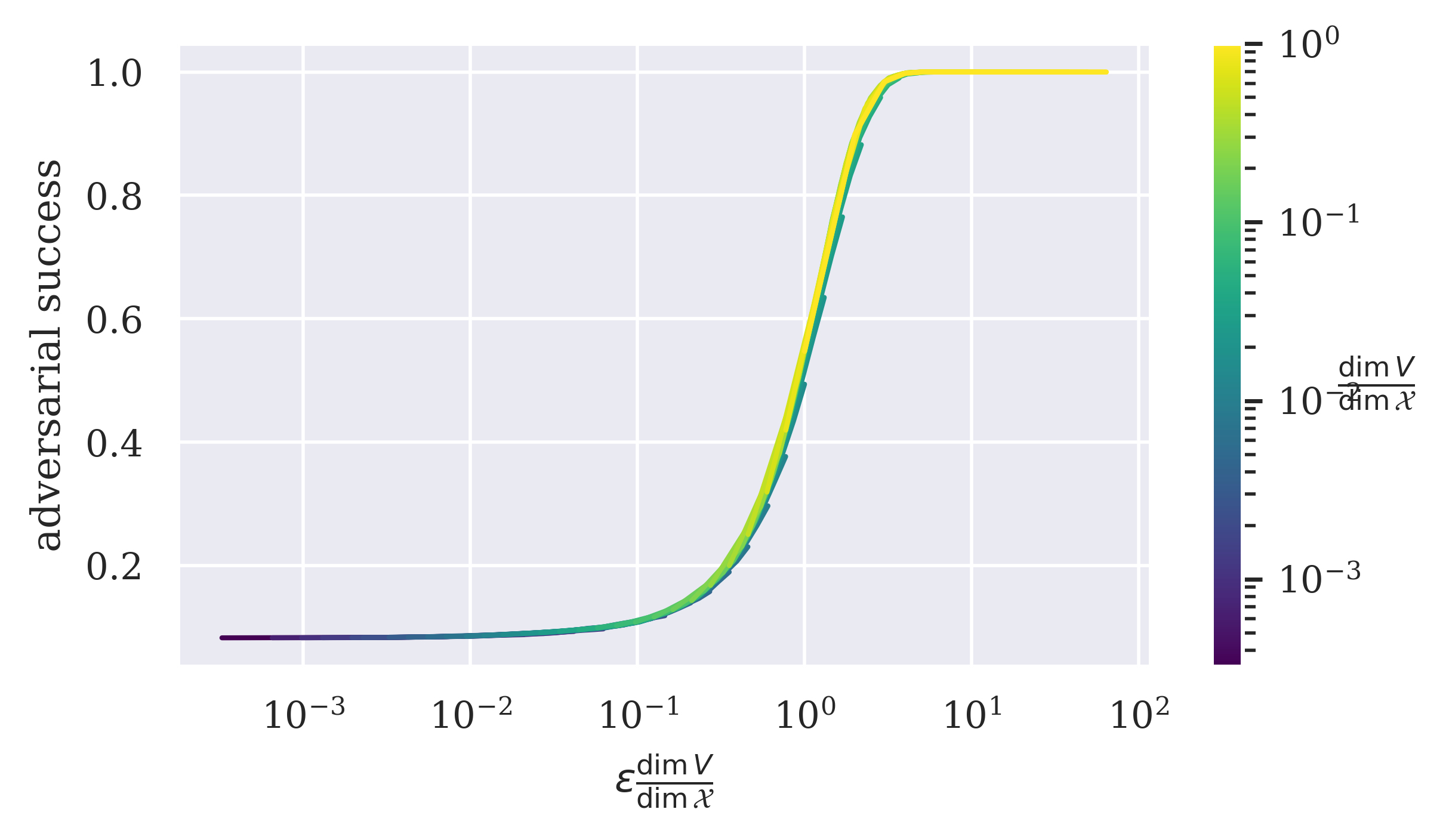}
      \caption[]{}\label{fig:infty-d}
   \end{subfigure}
   \caption{Plot for experiments analogous to those found in Figure \ref{fig:infty} but run with a ResNet9 trained and evaluated on CIFAR10.}
   \label{fig:infty-CIFAR}
\end{figure*}

\begin{figure*}[tb]
    \centering
   \begin{subfigure}{0.45\linewidth}
      \centering
      \includegraphics[width=\linewidth]{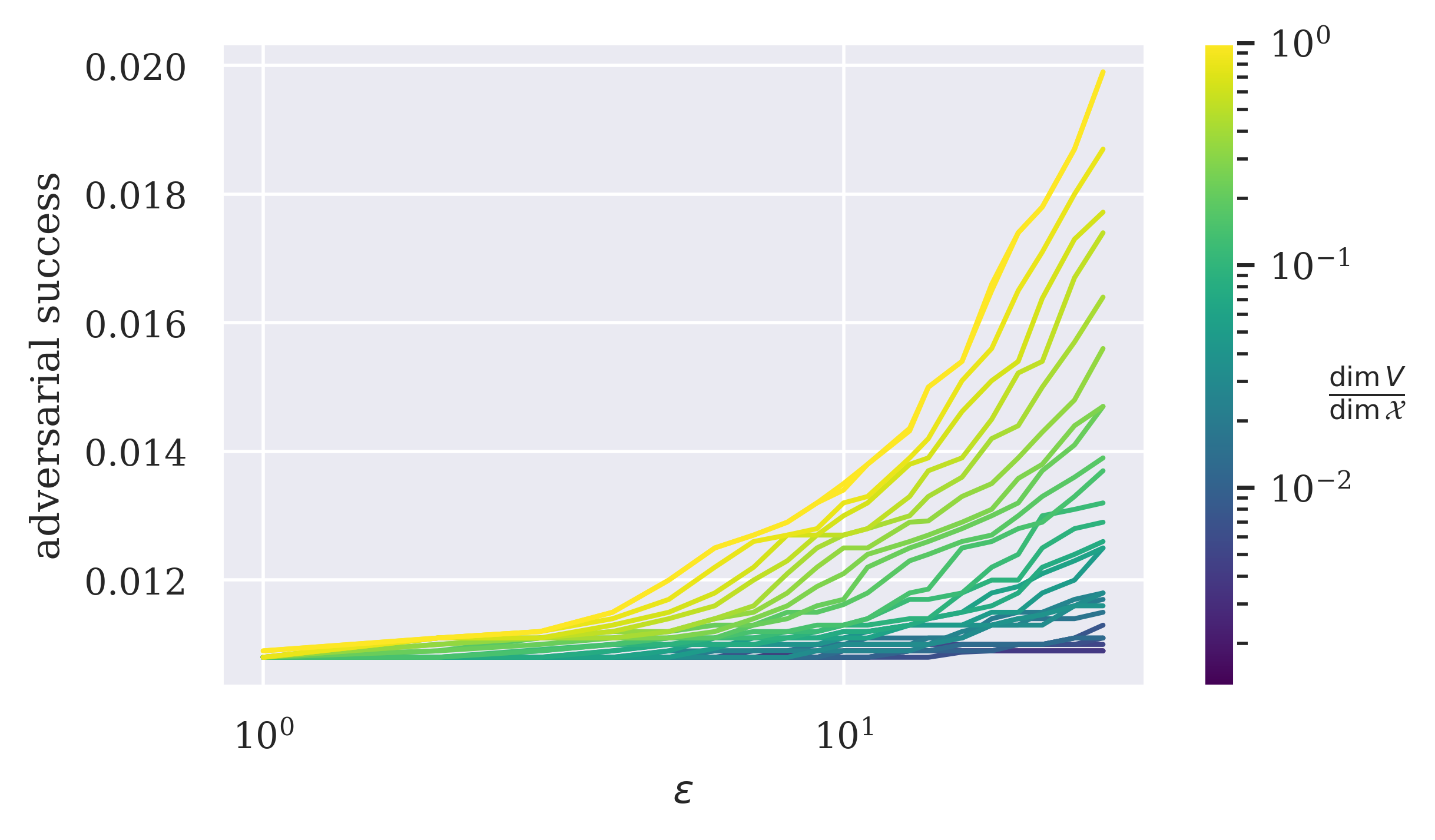}
      \caption[]{}\label{fig:2-a}
   \end{subfigure}
   \begin{subfigure}{0.45\linewidth}
      \centering
      \includegraphics[width=\linewidth]{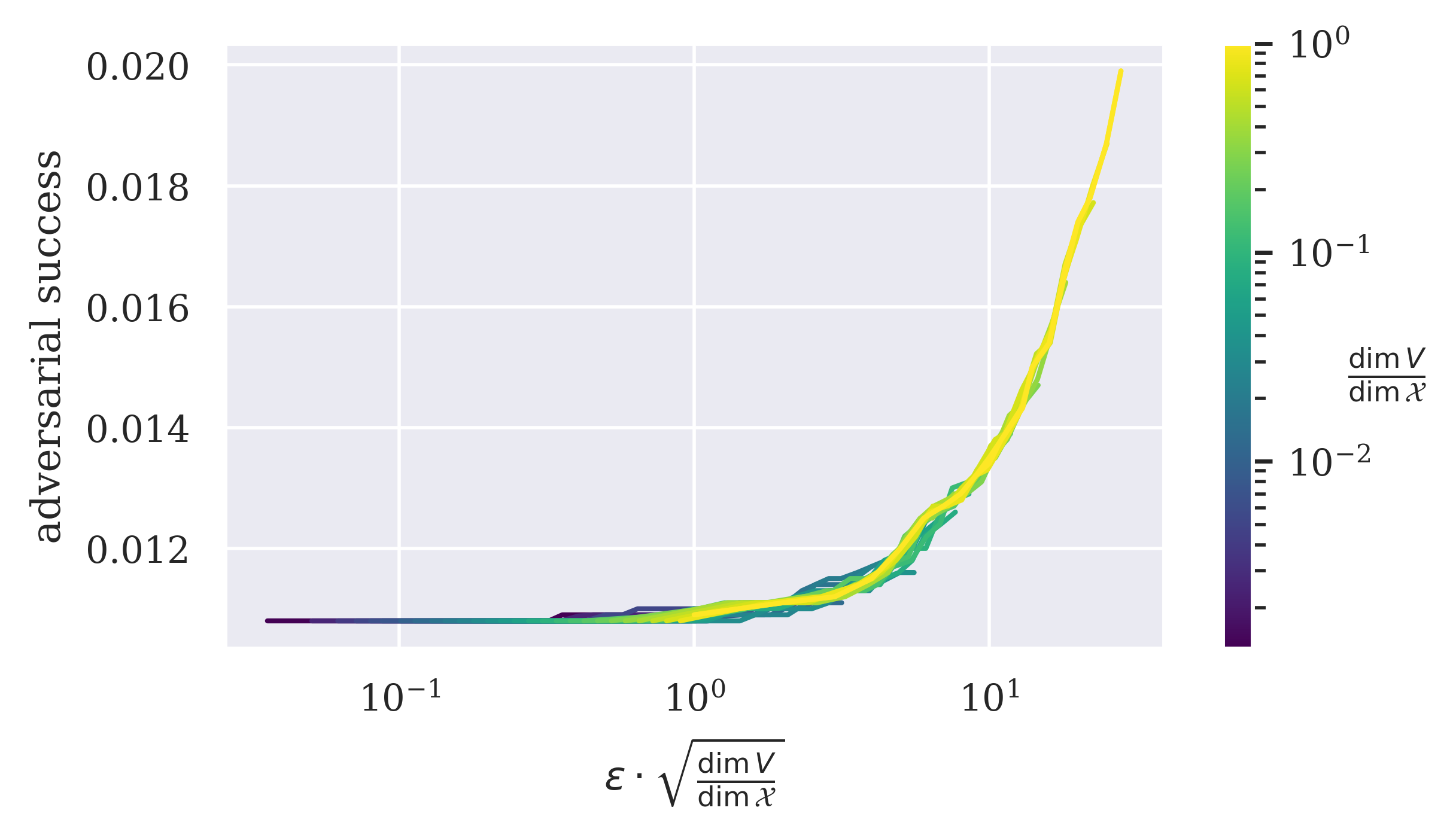}
      \caption[]{}\label{fig:2-b}
   \end{subfigure}
   \caption[]{Plot for experiments analogous to those found in Figure \ref{fig-L2-imagenet} but run with a 2-layer CNN trained and evaluated on MNIST.}
   \label{fig-L2-MNIST}
\end{figure*}

\begin{figure*}[tb]
    \centering
   \begin{subfigure}{0.45\linewidth}
      \centering
      \includegraphics[width=\linewidth]{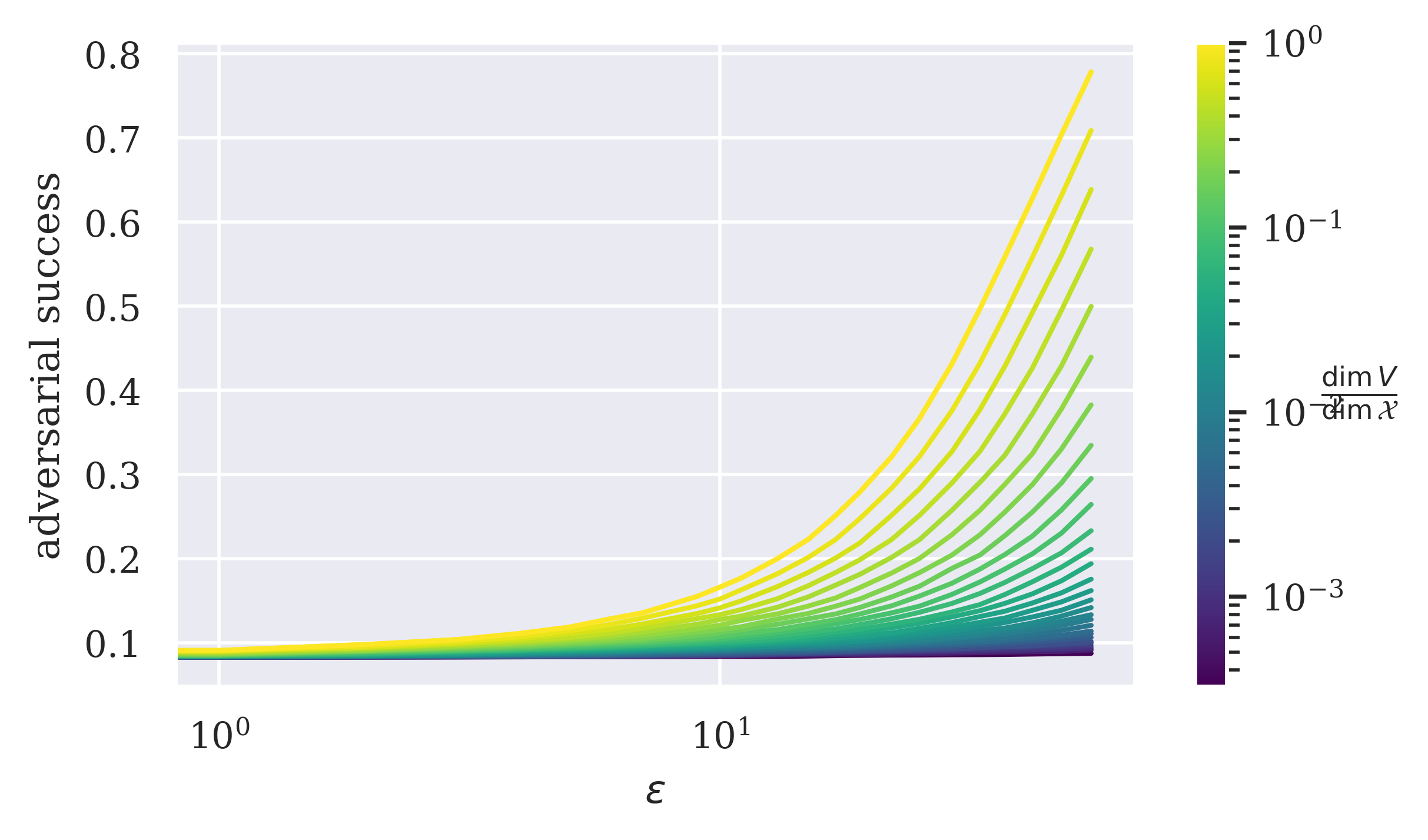}
      \caption[]{}\label{fig:2-c}
   \end{subfigure}
   \begin{subfigure}{0.45\linewidth}
      \centering
      \includegraphics[width=\linewidth]{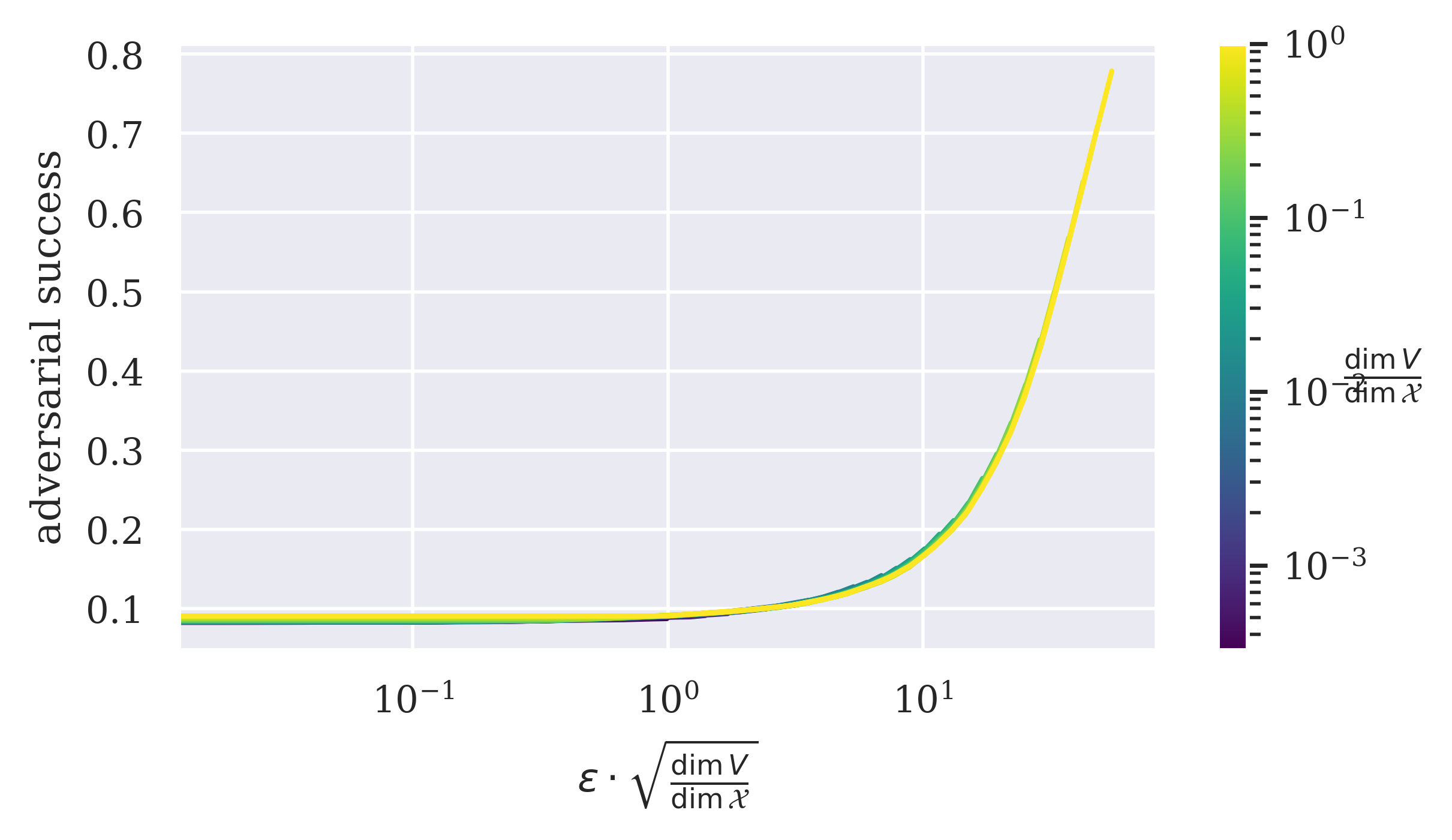}
      \caption[]{}\label{fig:2-d}
   \end{subfigure}
   \caption[]{
   Plot for experiments analogous to those found in Figure \ref{fig-L2-imagenet} but run with a ResNet9 trained and evaluated on CIFAR10.}
   \label{fig-L2-CIFAR10}
\end{figure*}

\begin{figure*}[tb]
    \centering
   \begin{subfigure}{0.45\linewidth}
        \centering
        \includegraphics[width=\linewidth]{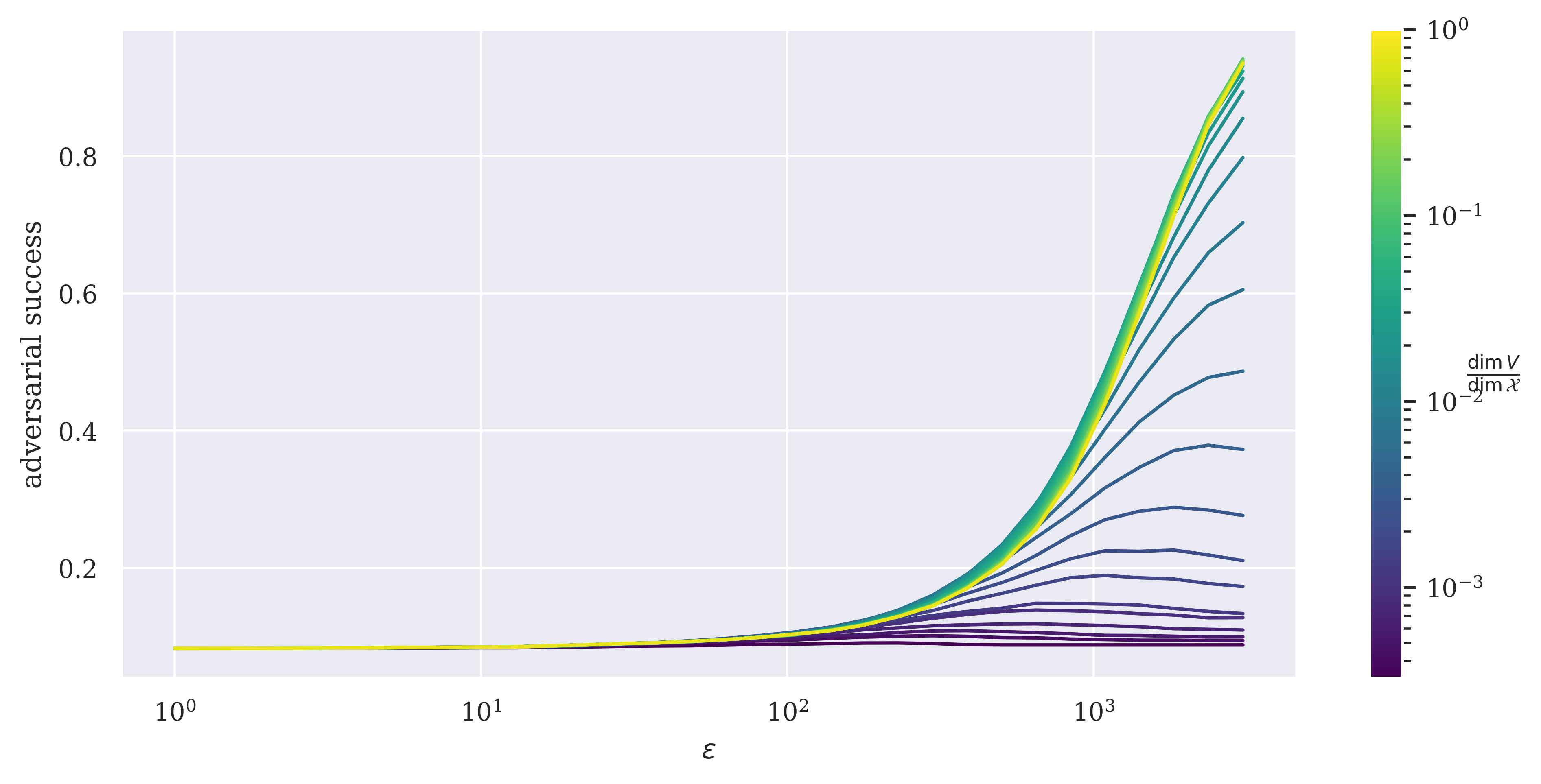}
        \caption[]{}\label{fig:cifar-1-a}
    \end{subfigure}
    \begin{subfigure}{0.45\linewidth}
        \centering
        \includegraphics[width=\linewidth]{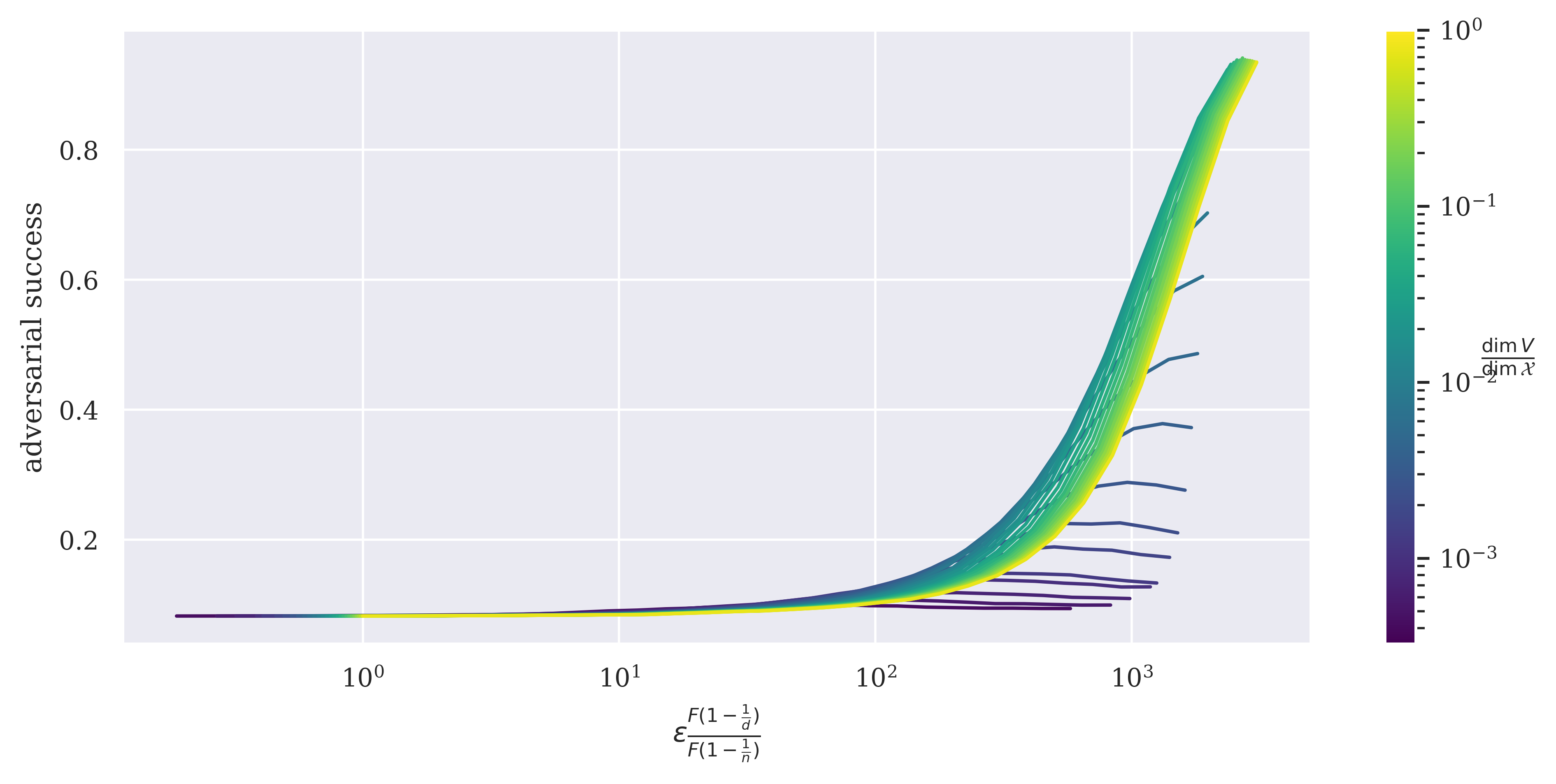}
        \caption[]{}\label{fig:cifar-1-b}
    \end{subfigure}
    \caption{Plot for experiments analogous to those found in Figure \ref{fig-L1-MNIST} but run with a ResNet9 trained and evaluated on CIFAR10.}
   \label{fig-L1-cifar}
\end{figure*}

\begin{figure*}[tb]
    \centering
   \begin{subfigure}{0.45\linewidth}
        \centering
        \includegraphics[width=\linewidth]{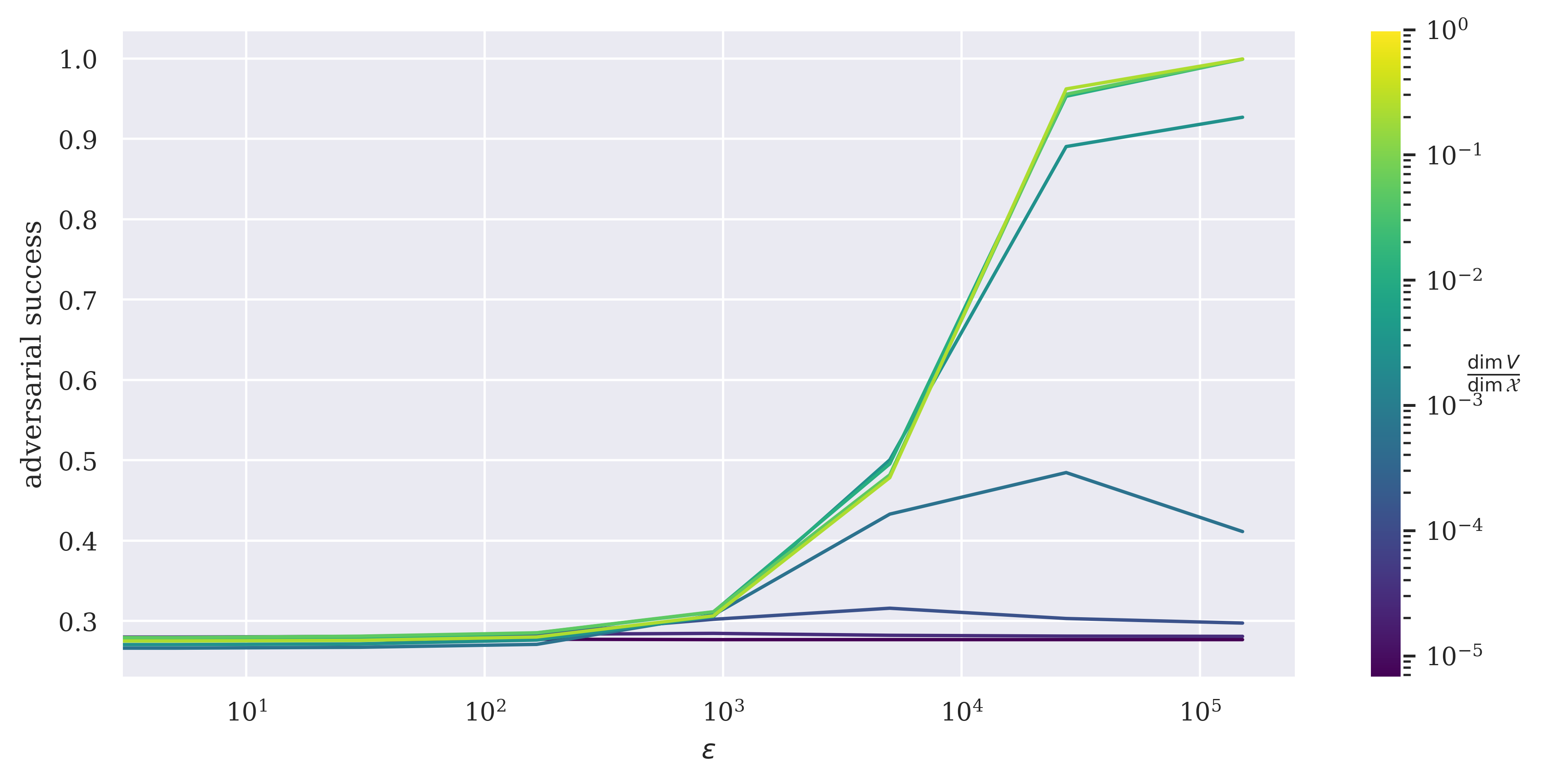}
        \caption[]{}\label{fig:1-e}
    \end{subfigure}
    \begin{subfigure}{0.45\linewidth}
        \centering
        \includegraphics[width=\linewidth]{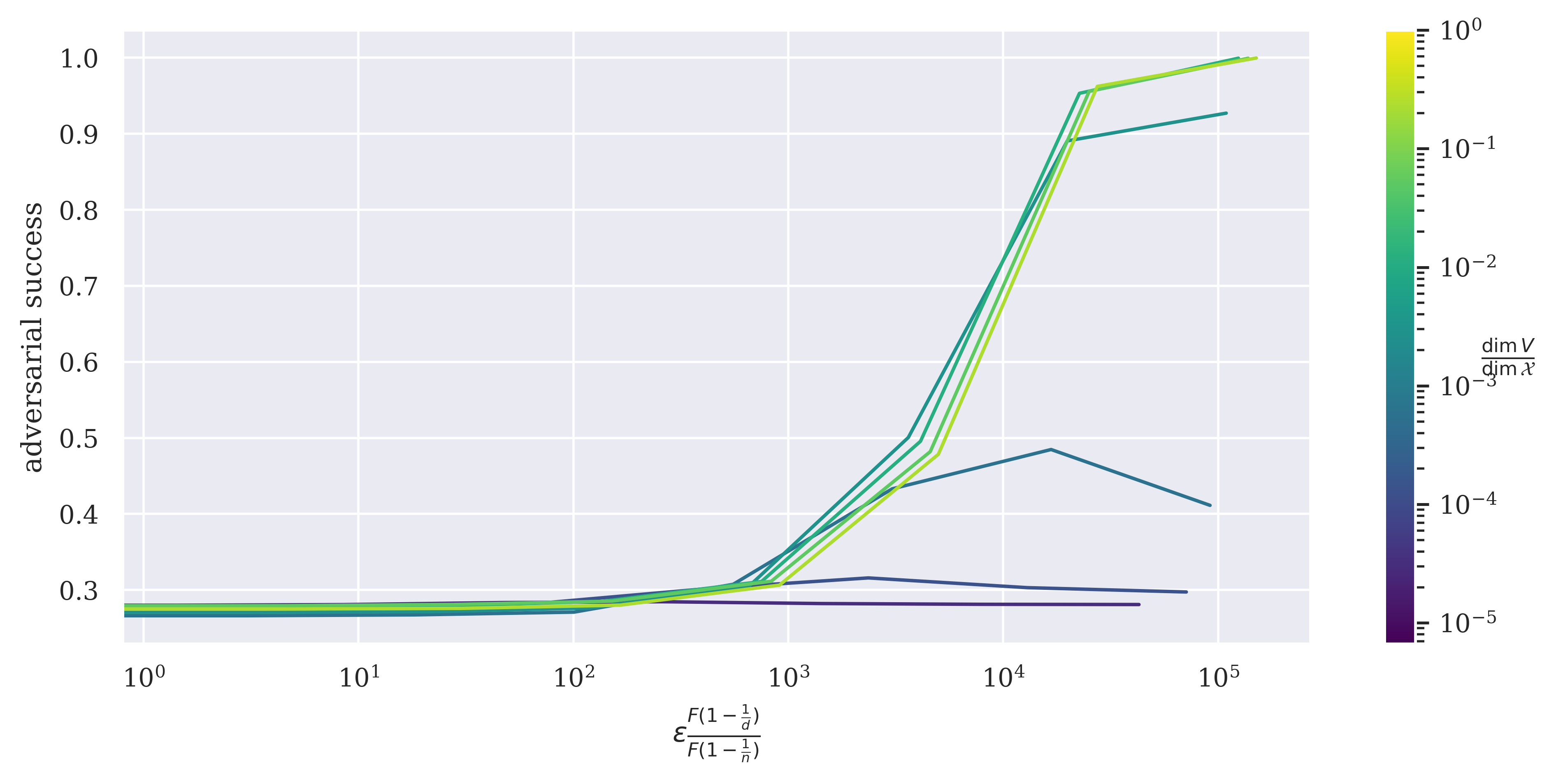}
        \caption[]{}\label{fig:1-f}
    \end{subfigure}
    \caption{Plot for experiments analogous to those found in \cref{fig-L1-MNIST} but run with a ResNet50 trained and evaluated on ImageNet.}
   \label{fig-L1-imagenet}
\end{figure*}

\begin{figure*}[tb]
    \centering
   \begin{subfigure}{0.45\linewidth}
        \centering
        \includegraphics[width=\linewidth]{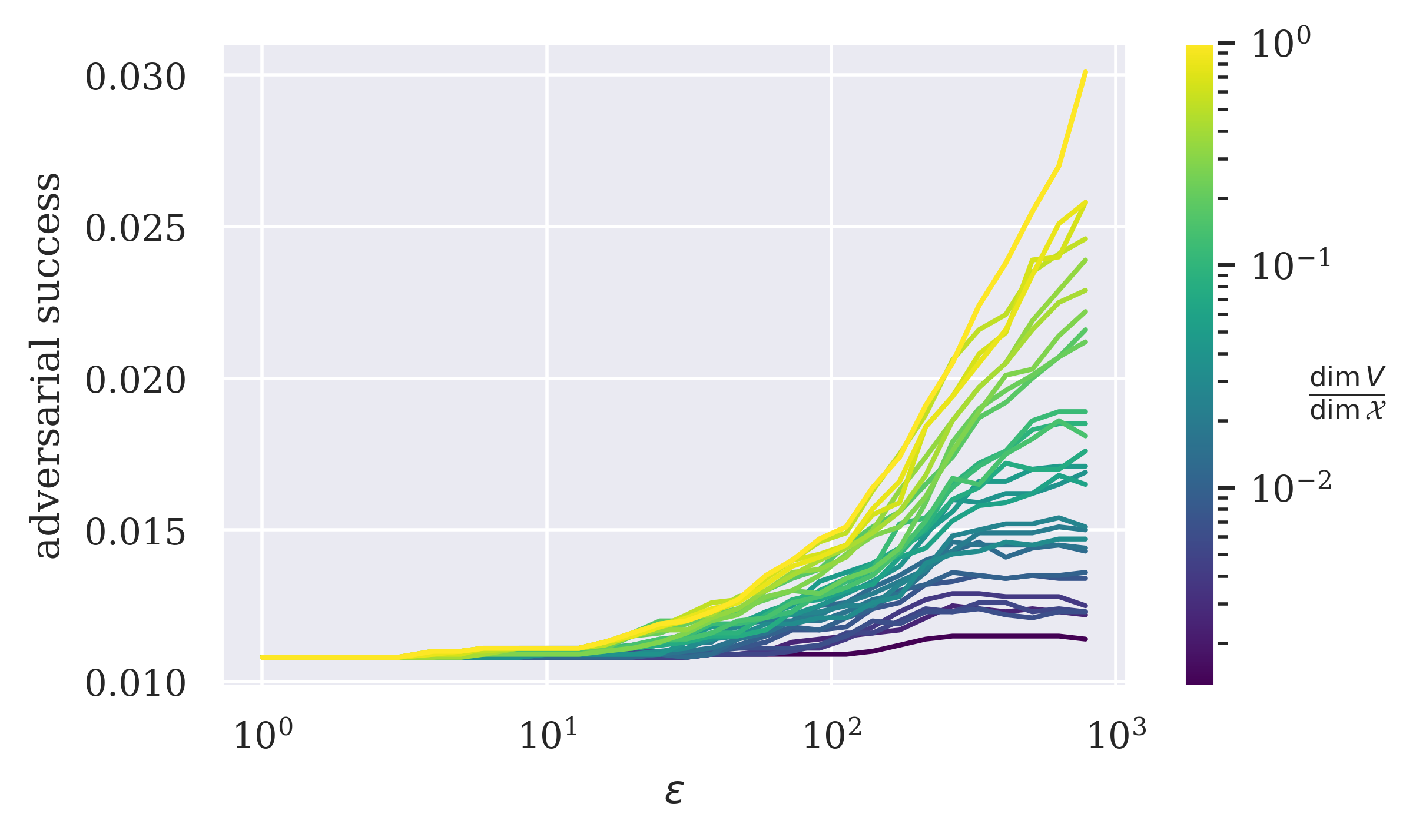}
        \caption[]{}\label{fig:mnist-1-a-new}
    \end{subfigure}
    \begin{subfigure}{0.45\linewidth}
        \centering
        \includegraphics[width=\linewidth]{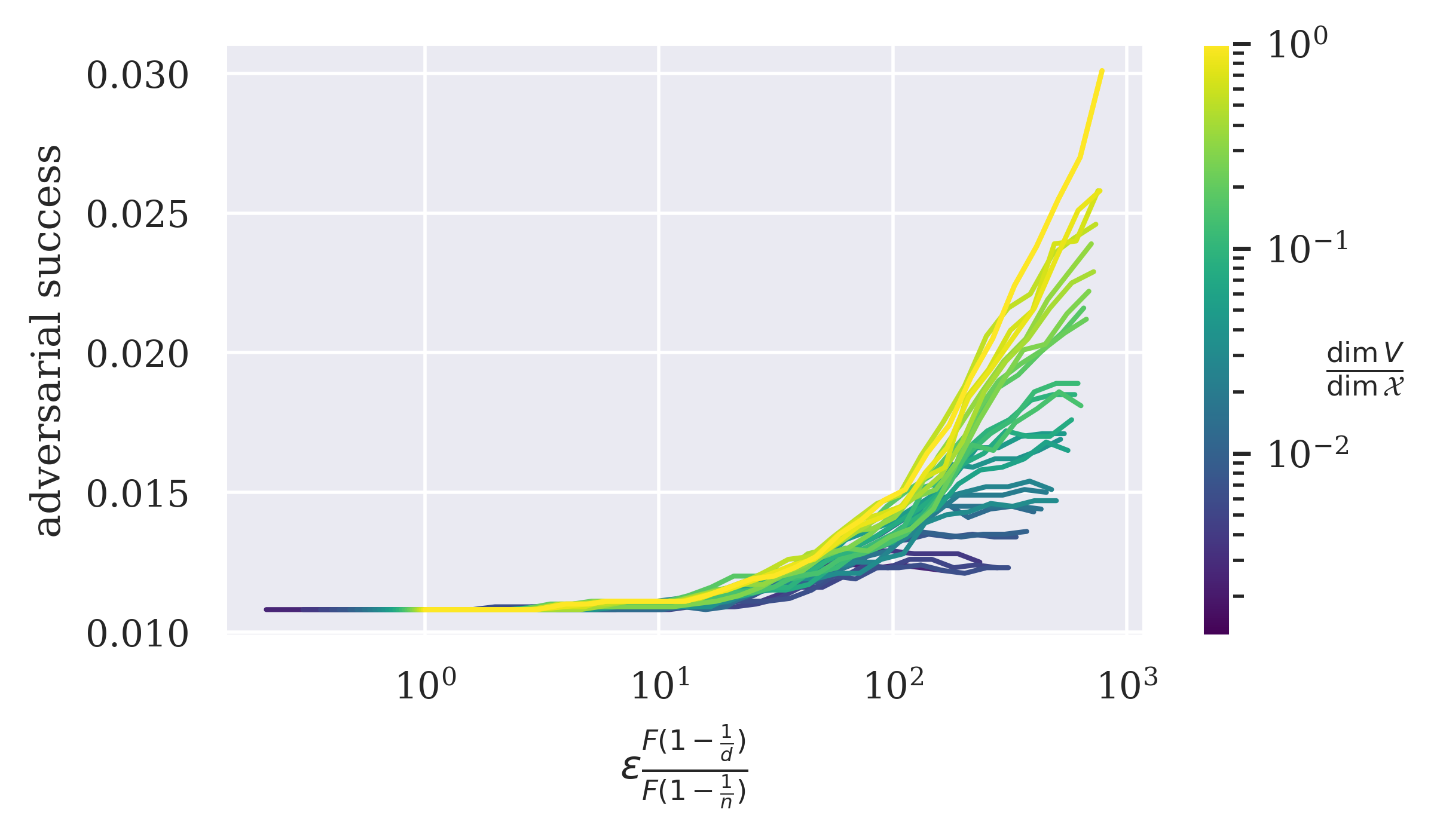}
        \caption[]{}\label{fig:mnist-1-b-new}
    \end{subfigure}
    \caption{Plot for experiments analogous to those found in Figure \ref{fig-L1-MNIST}, the only difference being that we use the FGSM step of \cref{eq:fgsm-1}.}
   \label{fig-L1-MNIST-new}
\end{figure*}

\begin{figure*}[tb]
    \centering
   \begin{subfigure}{0.45\linewidth}
        \centering
        \includegraphics[width=\linewidth]{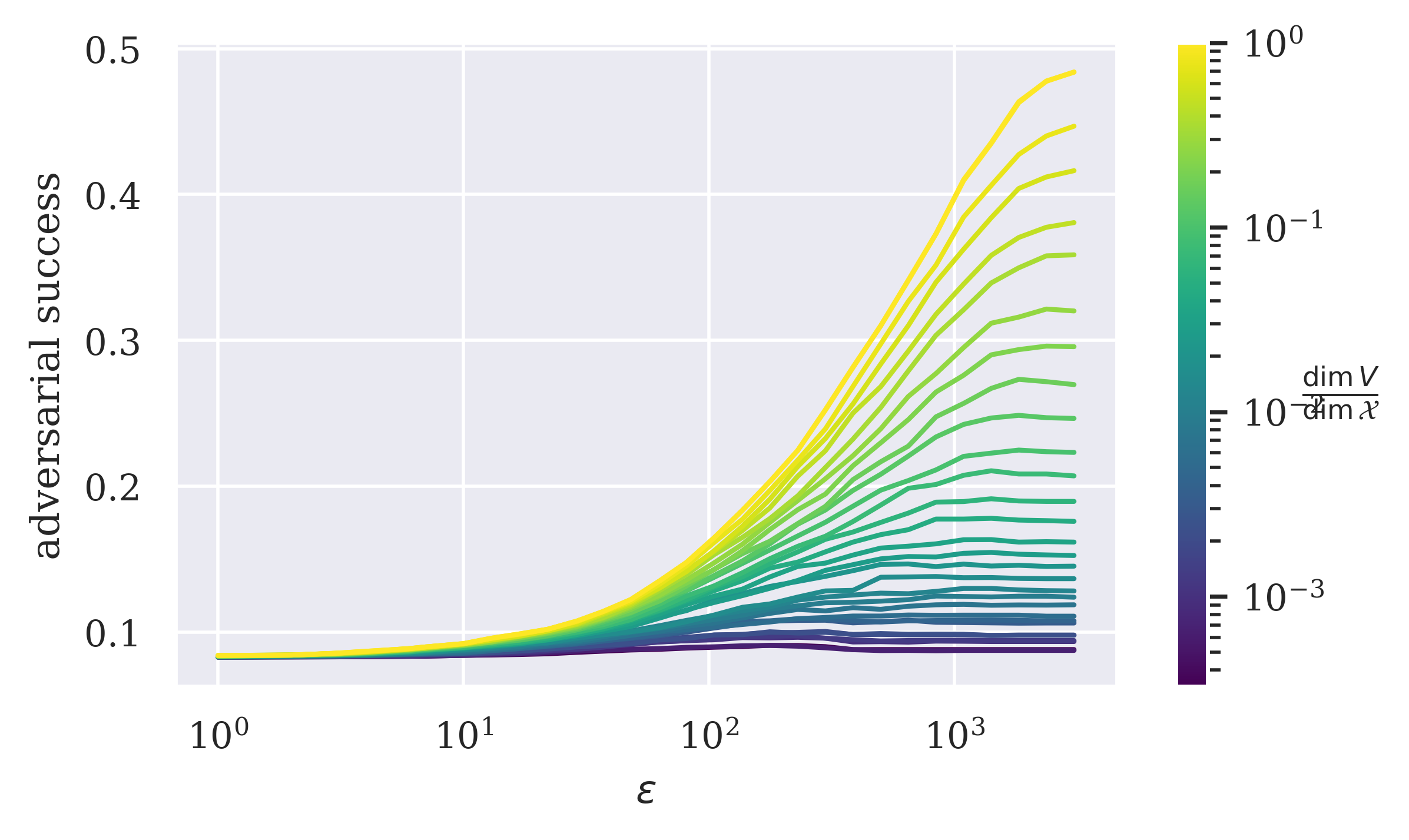}
        \caption[]{}\label{fig:cifar-1-a-new}
    \end{subfigure}
    \begin{subfigure}{0.45\linewidth}
        \centering
        \includegraphics[width=\linewidth]{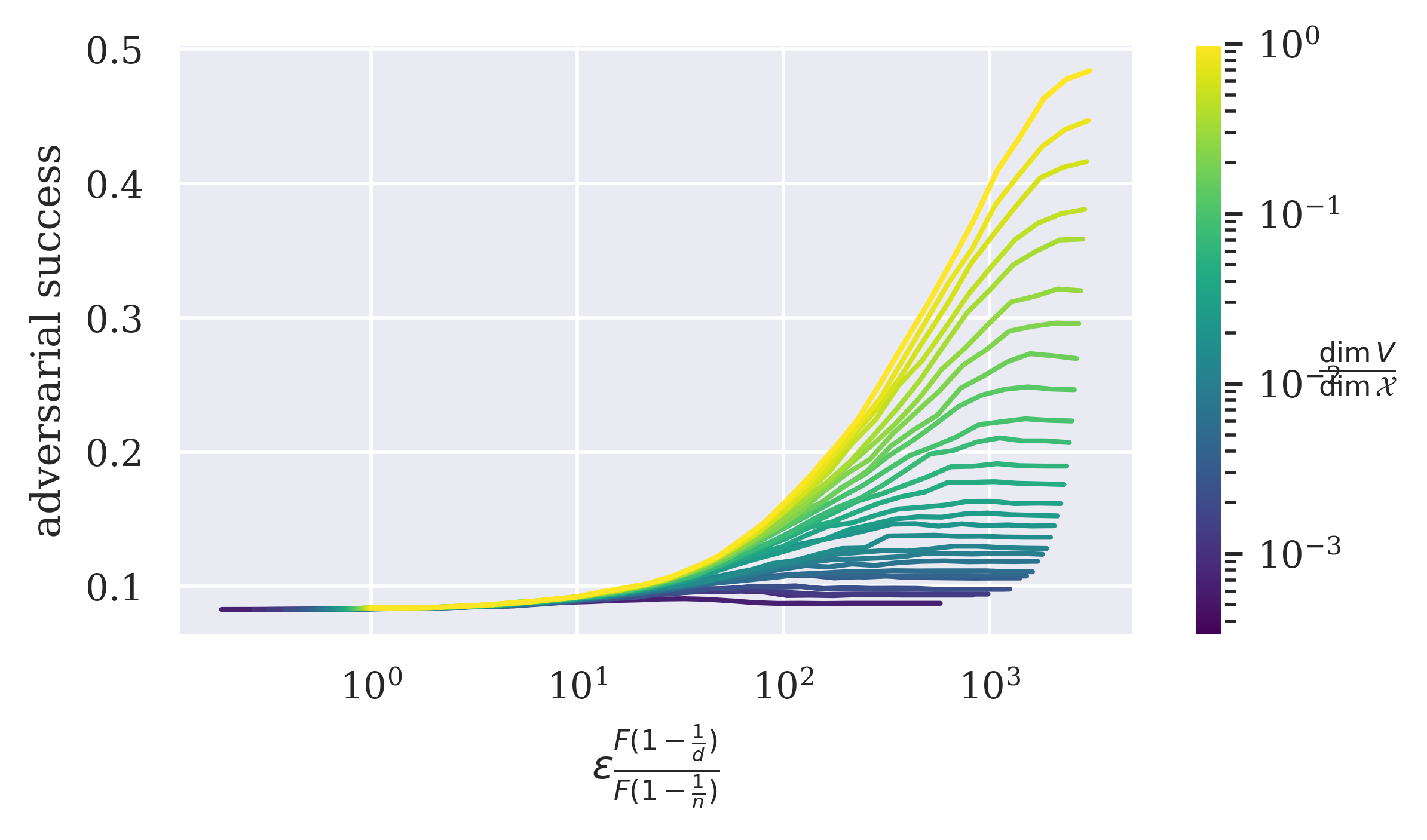}
        \caption[]{}\label{fig:cifar-1-b-new}
    \end{subfigure}
    \caption{Plot for experiments analogous to those found in Figure \ref{fig-L1-MNIST} but run with a ResNet9 trained and evaluated on CIFAR10, using the FGSM step of \cref{eq:fgsm-1}.}
   \label{fig-L1-cifar-new}
\end{figure*}

\begin{figure*}[tb]
    \centering
   \begin{subfigure}{0.45\linewidth}
        \centering
        \includegraphics[width=\linewidth]{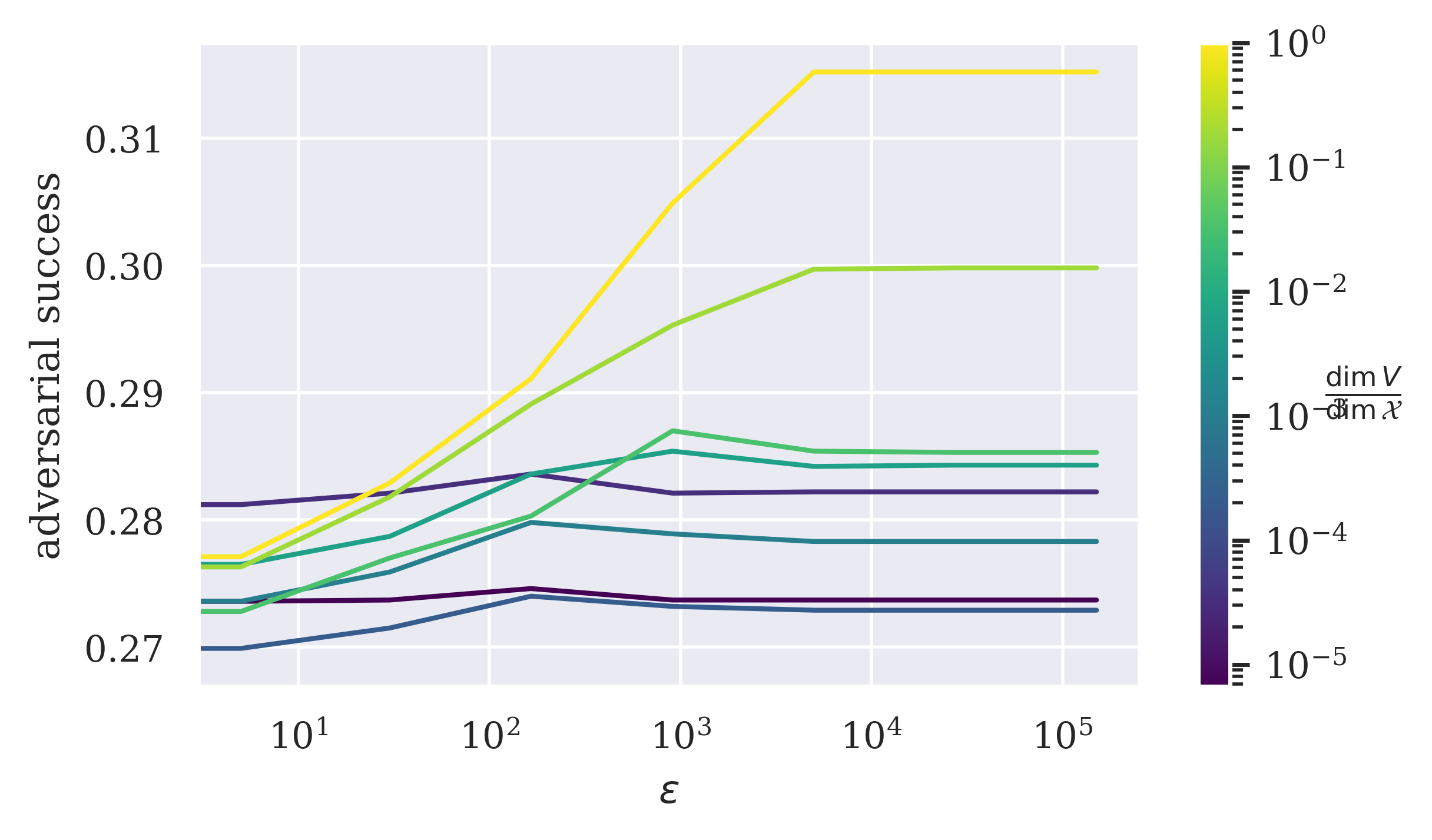}
        \caption[]{}\label{fig:1-e-new}
    \end{subfigure}
    \begin{subfigure}{0.45\linewidth}
        \centering
        \includegraphics[width=\linewidth]{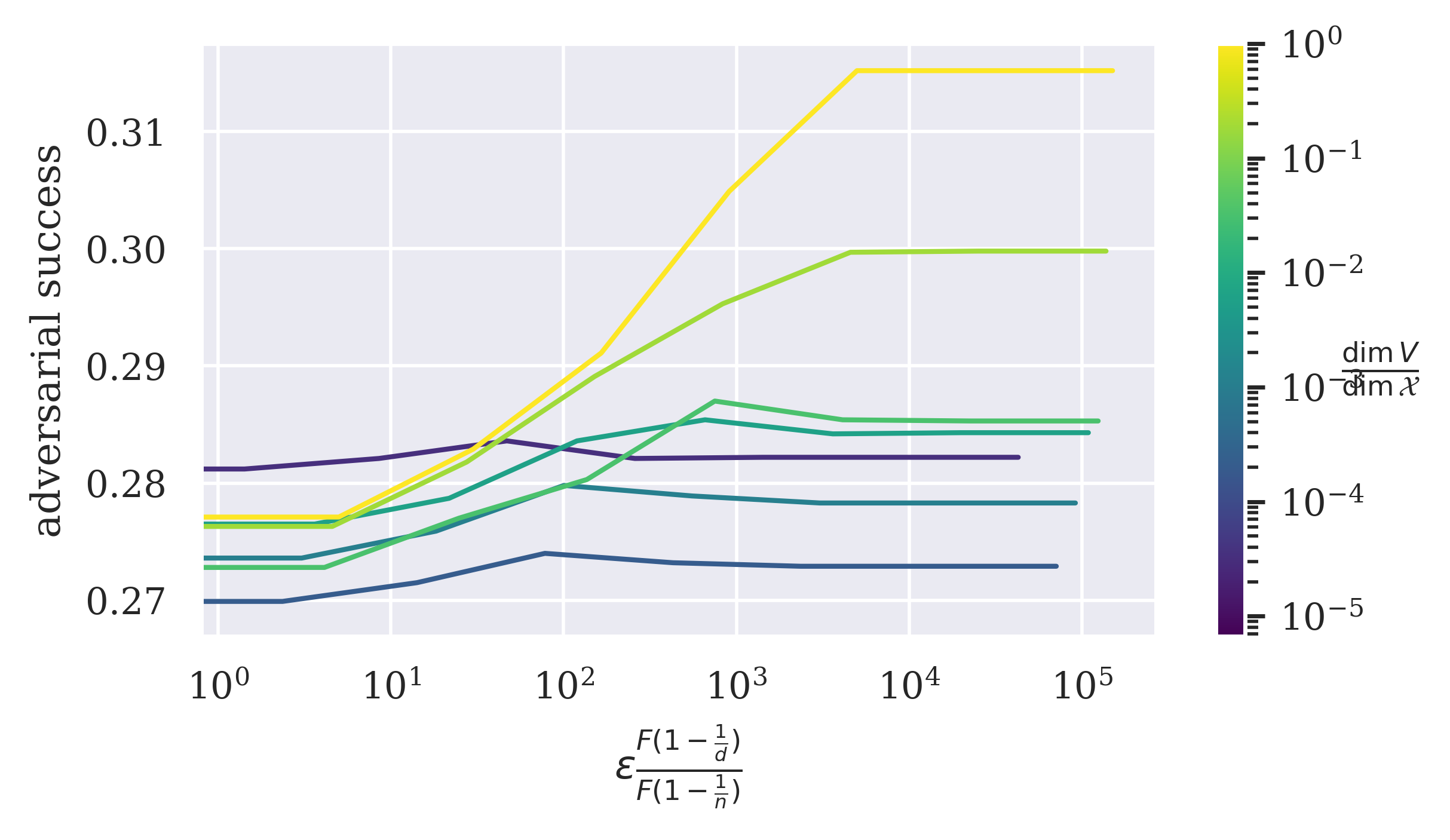}
        \caption[]{}\label{fig:1-f-new}
    \end{subfigure}
    \caption{Plot for experiments analogous to those found in \cref{fig-L1-MNIST} but run with a ResNet50 trained and evaluated on ImageNet, using the FGSM step of \cref{eq:fgsm-1}.}
   \label{fig-L1-imagenet-new}
\end{figure*}